\documentclass{article} % For LaTeX2e
\usepackage{colm2024_conference}

\usepackage{microtype}
\usepackage{hyperref}
\usepackage{url}
\usepackage{booktabs}
\usepackage{paralist}
\usepackage{soul}
\usepackage{graphicx}
\usepackage{amsmath}
\usepackage{amsthm}
\usepackage{algorithm}
\usepackage{algorithmic}
\usepackage[switch]{lineno}
\usepackage{multirow}
\usepackage{multicol}
\usepackage{wrapfig}

\usepackage{amssymb}
\usepackage{mathtools}
\usepackage{color}
\usepackage{colortbl}
\usepackage{tcolorbox}
\usepackage{enumitem}
\usepackage{listings}
\usepackage{makecell}
\usepackage{pifont}
\usepackage{subcaption}

\usepackage{tikz}
\usepackage{mdframed}
\gdef\Sepline{%
  \par\noindent\makebox[\linewidth][l]{%
  \hspace*{-\mdflength{innerleftmargin}}%
   \tikz\draw[thick,dashed,gray!60] (0,0) --%
(\textwidth+\the\mdflength{innerleftmargin}+\the\mdflength{innerrightmargin},0);
  }\par\nobreak}

\usepackage{xcolor}
\definecolor{msred}{RGB}{237,27,36}
\definecolor{msgreen}{RGB}{0,163,0}
\definecolor{msblue}{RGB}{0,114,198}
\definecolor{msyellow}{RGB}{255,187,0}
\definecolor{darkgreen}{RGB}{50,100,0}
\definecolor{darkred}{RGB}{200, 0, 0}
\newcommand{\cmark}{\textcolor{darkgreen}{\ding{51}}} %
\newcommand{\xmark}{\textcolor{darkred}{\ding{55}}} %

\definecolor{lightblue}{RGB}{210, 220, 250}

\usepackage{amsmath,amsfonts,bm}

\def\eqref#1{equation~\ref{#1}}
\def\1{\bm{1}}

\DeclareMathAlphabet{\mathsfit}{\encodingdefault}{\sfdefault}{m}{sl}
\SetMathAlphabet{\mathsfit}{bold}{\encodingdefault}{\sfdefault}{bx}{n}

\usepackage[noabbrev,capitalize]{cleveref}

\crefname{equation}{equation}{equations}
\crefname{line}{line}{lines}
\crefname{section}{\S}{\S\S}
\Crefformat{section}{#2\S#1#3}
\crefrangeformat{section}{\S\S#3#1#4--#5#2#6}
\title{Key-Point-Driven Data Synthesis with its Enhancement on Mathematical Reasoning}
\author{Yiming Huang\thanks{This work was done during the internship of Yiming Huang at Microsoft.},\quad
Xiao Liu\thanks{Corresponding author.},\quad
Yeyun Gong,\quad
Zhibin Gou,\quad
Yelong Shen,\\
\textbf{Nan Duan,\quad
Weizhu Chen}
\\
Microsoft
\\
\tt{\{v-yimhuang, xiaoliu2, yegong, v-zhibingou, yeshe\}@microsoft.com} \\
\tt{\{nanduan, wzchen\}@microsoft.com}
}

\colmfinalcopy % Uncomment for camera-ready version, but NOT for submission.
\begin{document}

\maketitle

\begin{abstract}
Large language models (LLMs) have shown great potential in complex reasoning tasks, yet their performance is often hampered by the scarcity of high-quality and reasoning-focused training datasets. Addressing this challenge, we propose Key-Point-Driven Data Synthesis (KPDDS), a novel data synthesis framework that synthesizes question-answer pairs by leveraging key points and exemplar practices from authentic data sources. KPDDS ensures the generation of novel questions with rigorous quality control and substantial scalability.
As a result, we present KPMath, an extensive synthetic dataset tailored for mathematical reasoning, comprising over 800K question-answer pairs.
Utilizing KPMath and augmenting it with additional reasoning-intensive corpora, we create the comprehensive KPMath-Plus dataset. 
% KPMath-Plus has significantly improved performance by fine-tuning multiple models.
% The fine-tuned DeepSeekMath model on Qwen1.5-72B achieves zero-shot PASS@1 accuracies of 87.0\% on GSM8K and 58.3\% on MATH, and also reaches promising performance on other math reasoning datasets, outperforming competitors in the 7B to 70B range.
The Qwen1.5-72B model, fine-tuned on KPMath-Plus, achieves 87.0\% PASS@1 accuracy on GSM8K and 58.3\% on MATH, surpassing competitors in the 7B to 70B range and best commercial models like GPT-4 across multiple math reasoning datasets.
% The fine-tuned Mistral-7B model on KPMath-Plus achieves a zero-shot PASS@1 accuracy of 82.1\% on GSM8K and 46.8\% on MATH, outperforming competitors in the 7B to 70B range. It achieves a promising performance on multiple math reasoning datasets.
% Our ablation studies further confirm the substantial enhancement in mathematical reasoning across various datasets, marking a significant stride in LLMs' reasoning capabilities.
\end{abstract}

\section{Introduction}
The recent advent of large language models (LLMs) such as GPT-4 \citep{openai2023gpt4}, Gemini \citep{team2023gemini}, and Mistral \citep{mistral2024large} has sparked significant interest due to their advanced capabilities in diverse domains \cite{bubeck2023sparks}. Despite this, their reasoning prowess, particularly in challenging domains like advanced mathematics \citep{lewkowycz2022solving}, competitive programming \citep{huang2023competition}, and integrated vision-language planning \citep{cen2024using}, remains under scrutiny. In current mathematical reasoning corpora, such as OpenWebMath \citep{paster2023openwebmath} and MathPile \citep{wang2023generative}, the vast internet-sourced data often suffers from poor quality and relevance to the subject matter. Conversely, manually annotated high-quality datasets like the MATH dataset \citep{hendrycksmath2021} are scarce and sometimes lack detailed reasoning steps.

Prior efforts to boost the mathematical reasoning capabilities of LLMs using synthetic data have primarily adopted two strategies.
The first strategy focuses on augmenting existing datasets. It involves question rephrasing \citep{yu2023metamath} or generating similar questions \citep{yu2023metamath,luo2023wizardmath,liu2024augmenting}.
However, the primary issue is that the generated questions are not only textually or conceptually similar but also uncontrollable in their variations.
The second strategy seeks to broaden the training dataset by generating new questions from established knowledge concepts. Knowledge bases are either compiled from online educational resources, such as Khan Academy's math courses \citep{huang2024mustard}, or synthesized from scratch using models like GPT-4 \citep{li2024synthetic}. 
However, these methods depend on constructed knowledge that might not align with the existing dataset's distributions and are difficult to comprehend without examples to illustrate the concepts.

% % \setlength{\columnsep}{0.1cm}
% \begin{figure}[ht]
% \begin{minipage}{0.4\textwidth}
%   \centering
%   \includegraphics[width=1.0\textwidth]{figs/dataset.pdf}
%   \caption{Comparative analysis of fine-tuning tesults on Mistral-7B using various datasets, illustrating Both the total number and the synthetic number of data samples.}
%   \label{fig:dataset}
% \end{minipage}
% \hfill
% \begin{minipage}{0.58\textwidth}
%   \centering
% \includegraphics[width=1.0\textwidth]{figs/flow_chart2.pdf}
%   \caption{Process of synthesizing KPMath data.}
%   \label{fig:flow_chart}
% \end{minipage}
% % \vspace{-1em}
% \end{figure}

% \begin{figure}[t]
%     \centering
%     \begin{subfigure}{0.4\textwidth}
%         \centering
%         \includegraphics[width=1.0\textwidth]{figs/dataset.pdf}
%         \caption{}
%         \label{fig:dataset}
%     \end{subfigure}
%     \hfill
%     \begin{subfigure}{0.58\textwidth}
%         \centering
%         \includegraphics[width=1.0\textwidth]{figs/flow_chart2.pdf}
%         \caption{}
%         \label{fig:flow_chart}
%     \end{subfigure}
%     \caption{(a) Comparative analysis of fine-tuning results on Mistral-7B using various datasets, illustrating both the total number and the synthetic number of data samples. (b) Process of synthesizing KPMath data.}
%     \label{fig:figures}
% \end{figure}

\begin{figure}[t]
\centering
\includegraphics[width=\textwidth]{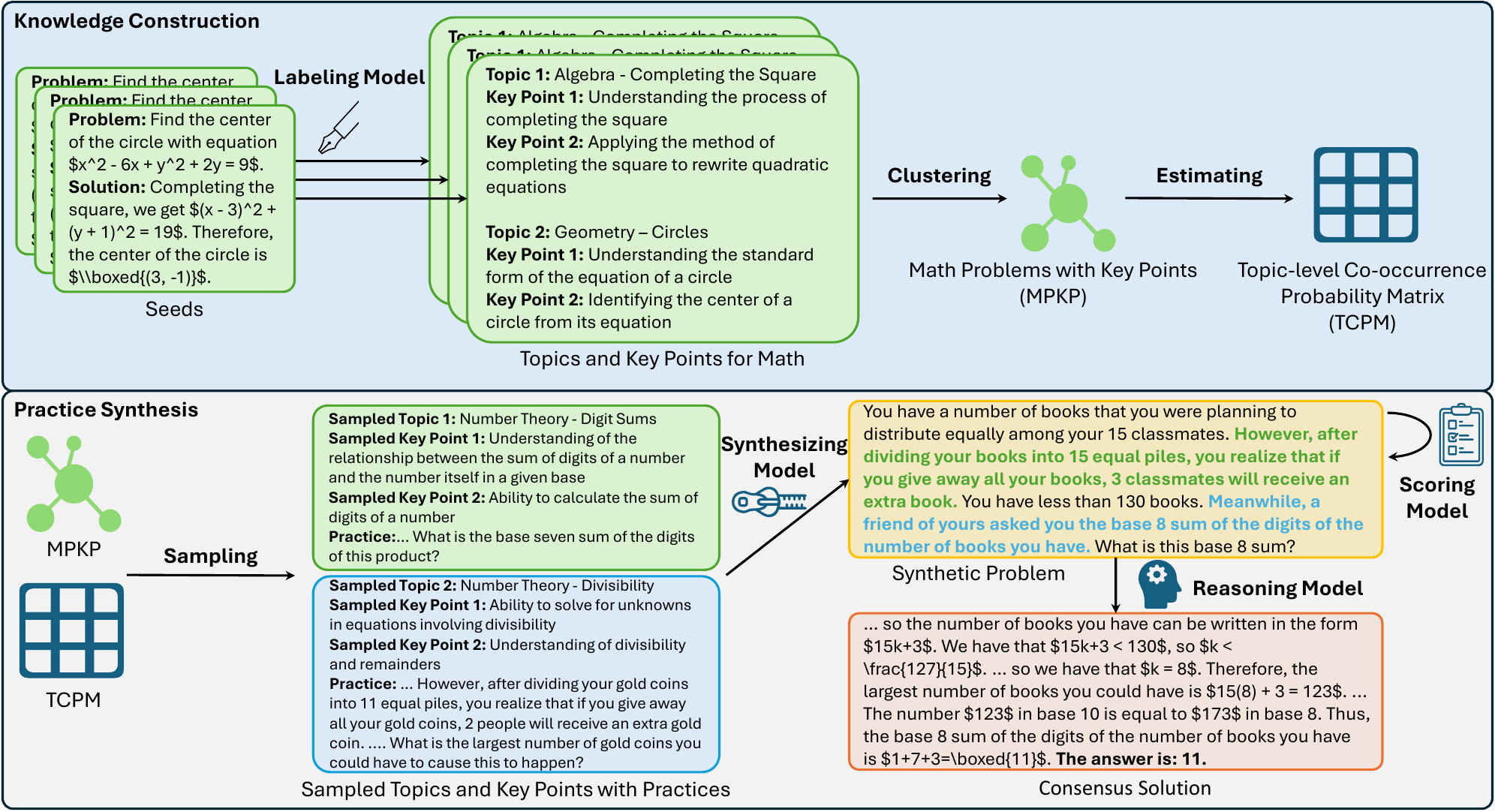}
\caption{Overview of the Key-Point-Driven Data Synthesis (KPDDS) pipeline, from knowledge extraction to practice synthesis.}
\label{fig:main}
\end{figure}

Considering these disadvantages of the two strategies, we introduce a novel data synthesis paradigm termed \textbf{Key-Point-Driven Data Synthesis (KPDDS)}, which capitalizes on the strengths of both data synthesis strategies. As depicted in \cref{fig:main}, it delves into datasets for knowledge mining, using relevant key points and associated problems to inform the generation of new problems.
(1) For knowledge construction, we begin by extracting topics and key points from seed problems using a labeling model, followed by a clustering algorithm to ensure deduplication and alignment.
Therefore, we get the Math Practices with Key Points (MPKP) dataset and construct the Topic-level Co-occurrence Probability Matrix (TCPM) to understand the frequency and distribution of topic pairs within the dataset.
(2) For practice synthesis, we sample multiple topics and key points from MPKP using the TCPM as a guide. These key points, along with corresponding example practices, serve as input for the synthesizing model to generate new questions. A scoring model then assesses the quality of these questions, allowing only those with high scores to proceed. Then, a reasoning model generates a range of answer options, which are later consolidated into consensus solutions through a voting mechanism.

Utilizing the training sets of the MATH~\citep{hendrycksmath2021} and GSM8K~\citep{GSM8K} datasets as foundational data, we developed a novel dataset named \textbf{KPMath}. Our training corpus was further enriched by integrating a series of mathematical reasoning datasets, leading to the creation of a comprehensive training dataset, \textbf{KPMath-Plus}. By fine-tuning the Qwen1.5-72 model \citep{bai2023qwen} on KPMath-Plus, we achieved zero-shot PASS@1 accuracies of 87.0\% on the GSM8K test set and 58.3\% on the MATH test set, culminating in a promising average of 81.5\% across six math reasoning datasets. This performance exceeds that of all competitors within the 7B to 70B model size range and best commercial models like GPT-4.
In the Hungarian Exam Score test, the KPMath-Plus-Mistral-7B model also outperforms the majority of models, indicating its competitive performance.

% For out-of-domain benchmarks, KPMath-Plus-Mistral-7B reaches optimal or sub-optimal performance on multiple datasets and attains state-of-the-art average performance on six mathematical reasoning tasks. 
%On the Hungarian Exam Score test, KPMath-Plus-Mistral-7B ranks just behind GPT-4~\citep{openai2023gpt4} and Grok-1~\citep{xai2023}, indicating its competitive performance.

% This level of performance not only surpasses that of other 7B models without coding usage but also exceeds the results of certain 34B models.

% In summary, our contributions include the following:
% \begin{itemize}
% \item We propose the KPDDS, a new paradigm of data synthesis. KPDDS leverages inferred topics and key points from authentic data, along with knowledge practices to guide the synthesis process while ensuring the question novelty, rigorous quality control, and dataset scalability.

% \item We develop KPMath, which is the largest synthetic dataset for mathematical reasoning to date. The Mistral-7B model, post fine-tuning with KPMath-Plus, outperforms 7B counterparts and exceeds select 34B models on the MATH test set.

% \item We introduce a dual quality control mechanism that employs self-scoring filtering and consensus voting, significantly improving dataset reliability. This demonstrates a straightforward yet effective approach to quality assurance in data synthesis.
% \end{itemize}

\section{Related Work}

\subsection{Math Reasoning with LLMs}

Recently, solving math problems is treated as an important aspect of evaluating LLM's reasoning ability.
However, the LLMs trained for general purposes like GPT-4 \citep{bubeck2023sparks}, Llama2 \citep{Touvron2023Llama2O}, Mistral \citep{jiang2023mistral}, InternLM2 \citep{2023internlm}, Qwen \citep{bai2023qwen}, Gemini \citep{team2023gemini} and DeepSeek \citep{bi2024deepseek} have shown limited capabilities in math reasoning.
To enhance the math reasoning ability of LLMs, researchers have turned their attention to research directions like prompting methods \citep{chia2023contrastive, zheng2023progressive, chen2023skills, zhang2023cumulative}, data construction for pretraining \citep{taylor2022galactica, lewkowycz2022solving, openwebmath, azerbayev2022proofpile, azerbayev2023llemma} and instruction tuning \citep{yue2023mammoth, yu2023metamath, luo2023wizardmath, gou2023tora, an2023learning, liu2024augmenting, huang2024mustard, li2024synthetic}, interacting with external tools \citep{mishra2022Lila, gao2022pal, gou2024critic, gou2023tora, yue2023mammoth,zhou2023solving, zhang2024evaluating}, and reinforcement learning with rewards \citep{ma2023let, yu2023outcome, wang2023math, luong2024reft} for either outcomes or steps.
This work is in line with math reasoning data construction for instruction tuning.

\subsection{Data Synthesis for Math Reasoning}
In the realm of math reasoning, data synthesis is usually applied for instruction tuning, with each data sample encompassing a question text and its corresponding answer text. To advance this field, research efforts focus on three critical aspects: enhancing the quality of answers, generating novel questions, and implementing quality control measures.
% For math reasoning, data synthesis is usually applied for instruction tuning, where each data sample can be viewed as a pair of a question text and an answer text.
% Apart from the question rephrasing and answer argumentation introduced in \cite{yu2023metamath}, the works on data synthesis for math reasoning can be introduced from three dimensions: question novelty, answer style, and quality control.

For answer quality, some works focus on chain-of-thought (CoT) \citep{wei2022chain,yu2023metamath} style answers, while others like \cite{yue2023mammoth} and \cite{gou2023tora} investigate program-based answers.
\cite{yue2023mammoth} synthesize program-of-thought (PoT) \citep{chen2022program} style answers using GPT-4.
\cite{gou2023tora} further explore interleaved answers with program-based tool use.
In this work, we focus on the synthesis of CoT-style answers.

For question novelty, research diverges into two approaches: starting from existing problems, \cite{shao2023synthetic} explore answer-first data synthesis and \cite{yu2023metamath} utilize backward reasoning, while \cite{luo2023wizardmath}, \cite{an2023learning}, and \cite{liu2024augmenting} focus on evolution instruction and iterative composition using reasoning steps. Alternatively, some work begins with knowledge-based techniques, where \cite{huang2024mustard} extracts concepts from Khan Academy and \cite{li2024synthetic} uses GPT-4 to create a concepts taxonomy. The former is limited by poor scalability with existing data, and the latter often yields a synthetic data distribution that significantly deviates from real data. In our work, we create questions by extracting key points from real data and then synthesizing new problems based on these key points with authentic and reliable exercises.

For synthetic data quality, \cite{huang2024mustard} prompt GPT-4 to convert CoT-style answers into verifiable Lean-3 code, while \citet{trinh2024solving}'s AlphaGeometry ensures Euclidean geometry theorem accuracy using symbolic deduction. In contrast, We assess synthetic question and answer quality through GPT-4 scored evaluations and consensus scoring via repeated sampling.

\subsection{Data Synthesis for Other Applications}

The aim of synthetic data is to offer a convincing and fuller depiction of the actual data source, maintaining key statistical characteristics such as the distribution patterns of continuous variables, categorical ratios, and the latent relationships among different variables.
Except for math reasoning, there are also works on data synthesis for other applications like code \citep{wizardcode, gunasekar2023textbooks, wei2023magicoder}, table reasoning \citep{lei2023s3eval}, medical application \citep{zhang2023alpacare, tang2023does}, visual reasoning \citep{du2023makes}, and general purposes \citep{wang2022self, xu2023wizardlm, li2024synthetic}.

\section{Method}

\subsection{Overview}

% Mathematical problems are intrinsically associated with specific knowledge domains. The range and complexity of required knowledge domains expand as the problems become more complex. Therefore, employing knowledge domains as a basis for question generation is a logical approach.
% MUSTARD \citep{huang2024mustard} has developed a repository of mathematical concepts from Khan Academy's math courses. However, these concepts are generally categorized under general terms such as ``Numbers and Operations'' and ``Foundations''. Additionally, the lack of exemplar problems in MUSTARD makes it difficult to generate well-formed mathematical questions from these overarching concepts. Furthermore, Khan Academy's grade-level categorization hinders the creation of complex problems that span multiple mathematical domains.

% Previous data synthesis efforts have either augmented original dataset with limited variation or relied on generic mathematical concepts, resulting in poor reliability in the generated data. Our approach synergizes these elements by extracting and leveraging knowledge from existing datasets to enhance both the diversity and reliability of the synthetic data. 
% In the comprehensive framework illustrated in \cref{fig:main}, our methodology is systematically delineated into three primary phases: Knowledge Extraction, Topic-level Co-occurrence Probability Matrix (TCPM) Construction, and Practice Generation. Each of these components will be elucidated in the subsequent sections. The specific prompts utilized for each component are detailed in \cref{sec:prompts}.
In the comprehensive framework illustrated in \cref{fig:main}, our methodology is systematically delineated into two primary phases: Knowledge Construction and Practice Generation, each consisting of two components.
We will introduce these four components separately: Knowledge Extraction, Topic-level Co-occurrence Probability Matrix (TCPM) Construction, Question Generation with Quality Assessment, and Answer Generation with Consensus Assessment.
The specific prompts utilized for each component are detailed in \cref{sec:prompts}.

\subsection{Knowledge Extraction}
\label{sec:knowledge_extraction}
We employ GPT-4 as the labeling model to extract knowledge pertinent to problem-solving from seed problems, as illustrated in \cref{fig:main}. 
The questions and solutions of seeds are input into GPT-4, which then extracts information at two levels of knowledge.
Key excerpts from the prompt for knowledge extraction are showcased in \cref{fig:label_prompt}, and the complete details are shown in \cref{fig:prompt_knowledge}.
The first level of knowledge is the topics, which correspond to the subject and its subcategories that are pertinent to the problem, such as "Geometry - Circles". 
The secondary level is key points (KPs), which comprise the theorems or methods essential for the resolution process, like "Determining the center of a circle from its equation". 

The process of knowledge extraction results in an uncontrolled, extensive number of topics, many of which exhibit semantic overlap. Examples of such redundancy include "Arithmetic - Percentages" and "Arithmetic - Percentage". Furthermore, there are instances where a topic occurs only once, accompanied by very few KPs. Therefore, we further process the extracted knowledge data. Specifically, we use OpenAI's text-embedding-ada-002 to embed all KPs, and the topics are represented by the average value of the embeddings of their included KPs. Then, we calculate the cosine similarity of the topic embeddings for deduplication and clustering, obtaining several representative topics, which are displayed in \cref{tab:topics_gsm,tab:topics_math}. Finally, we construct the Math Practices with Key Points (MPKP) dataset.

\subsection{TCPM Construction}
Mathematical problems typically involve multiple topics and KPs, and the combination of topics within these problems follows a discernible pattern. For example, semantically highly similar topics do not appear repeatedly in the same problem, whereas arbitrarily meshing unrelated topics tends to result in nonsensical questions. 
In light of this structured complexity, we compute the Topic-level Co-occurrence Probability Matrix (TCPM) from the topics present in mathematical questions within the MPKP dataset. Our methodology is systematically outlined in \cref{alg:co_occurrence}. This algorithm quantifies the co-occurrence and self-interaction of topics within a dataset by constructing a matrix that logs the frequency of topic pairs and the instances where the number of KPs for individual topics exceeds five, followed by a logarithmic normalization.
An increased co-occurrence probability between topic clusters indicates a likelihood of their concurrent appearance in the examined problems.
\cref{fig:co-occurrence_matrix_gsm,fig:co-occurrence_matrix_math} presents a heatmap visualization of the co-occurrence probability matrix.

% \begin{figure}[h]
\noindent
\begin{minipage}[t]{0.48\textwidth}
\vspace{0pt}
\begin{algorithm}[H]
    \small
    \caption{TCPM Calculation}
    \label{alg:co_occurrence}
    \begin{algorithmic}[1]
    \REQUIRE MPKP $dataset$
    \STATE $N \gets$  number of topics in $data$
    \STATE Initialize $TCPM$ with zeros of shape $N \times N$
    \FOR{each $d$ in $data$}
        \FOR{each topic $i$ in $d$}
            \FOR{each topic $j$ in $d$}
                \IF{$i \neq j$}
                    \STATE $TCPM[i][j] += 1$
                \ENDIF
            \ENDFOR
            \IF{Number of KPs in topic $i > 5$}
                \STATE $TCPM[i][i] += 1$
            \ENDIF
        \ENDFOR
    \ENDFOR
    \STATE $TCPM \gets \log_{10}(TCPM + 1)$
    \RETURN $TCPM$
    \end{algorithmic}
\end{algorithm}
\end{minipage}
\hfill
\begin{minipage}[t]{0.48\textwidth}
\vspace{10pt}
\begin{tcolorbox}[colback=msblue!5!white,colframe=msblue!80!black]
% \footnotesize
\fontsize{8pt}{8pt}\selectfont
% \textbf{Prompt for Knowledge Extraction:}

As a mathematics education specialist, please analyze the topics and key points of the provided question and its answer. These analysis should serve as a guide for teachers to craft analogous problems and as focal learning objectives for students when approaching the problem. Be sure to avoid repetition of Key Points for clarity and conciseness. Specific requirements are as follows: 

1. Identify and categorize the main mathematical topics involved in the problem. If knowledge from non-mathematical fields is used, it is classified into Others - xxx, such as Others - Problem Context.

2. For each topic, enumerate the essential Key Points relevant to the problem. 

...
\end{tcolorbox}
\vspace{-1em}
\captionof{figure}{Key excerpts of the prompt for knowledge extraction.}
\label{fig:label_prompt}
\end{minipage}
% \end{figure}

% \begin{algorithm}[tb]
%     \caption{Co-occurrence Matrix Calculation}
%     \label{alg:co_occurrence}
%  \begin{algorithmic}[1]
%  \REQUIRE MPKPP $data$
%  \STATE Initialize $co\_occurrence\_matrix$ with zeros of shape $num\_topics \times num\_topics$
%  \FOR{each $d$ in $data$}
%      \STATE $topic\_kps \gets$ math topic ids of $d$
%      \FOR{each topic $i$ in $topic\_kps$}
%          \FOR{each topic $j$ in $topic\_kps$}
%              \IF{$i \neq j$}
%                  \STATE $co\_occurrence\_matrix[i][j] \gets co\_occurrence\_matrix[i][j] + 1$
%              \ENDIF
%          \ENDFOR
%          \IF{Number of KPs in topic $i > 5$}
%              \STATE $co\_occurrence\_matrix[i][i] \gets co\_occurrence\_matrix[i][i] + 1$
%          \ENDIF
%      \ENDFOR
%  \ENDFOR
%  \STATE $co\_occurrence\_matrix \gets \log_{10}(co\_occurrence\_matrix + 1)$
%  \RETURN $co\_occurrence\_matrix$
%  \end{algorithmic}
%  \end{algorithm}

% \input{figs/prompt_kps}

% \input{figs/labeling_example}

\subsection{Question Generation with Quality Assessment}

% By extracting knowledge and constructing the TCPM from the seed problems, we pave the way for generating new problems that are similar yet varied in nature, building upon their foundational elements. Leveraging the TCPM, we proceed to randomly sample two to three topics, from which we then randomly select two KPs per topic, each KP is accompanied with an associated problem as the example. This process yields a foundational information set as the basis for our problem generation. Employing GPT-4, we use this set to generate new problems, with the prompt presented in \cref{fig:prompt_gen_q}.

By extracting knowledge and constructing the TCPM from the seed problems, we pave the way for generating new problems that are similar yet varied in nature, building upon their foundational elements. Leveraging the TCPM, we perform probabilistic sampling of topics, with the probability calculation method as follows:
\[  
V_n =   
\begin{cases}  
    \sum_{j} \text{TCPM}_{ij}, & \text{if } n = 1, \\  
    \text{TCPM}_{T_i, \cdot}, & \text{if } n = 2, \\  
    \text{TCPM}_{T_{n-1}, \cdot} \circ \text{TCPM}_{T_{n-2}, \cdot}, & \text{if } n > 2,  
\end{cases}  
\]  
where \(V_n\) represents the vector used for probabilistic topic sampling, \(i\) and \(j\) are index variables, \(T_i\) denotes the \(i\)-th topic, and \(\text{TCPM}_{T_{n}, \cdot}\) denotes the \(n\)-th row vector in TCPM. \(\circ\) denotes the Hadamard product (element-wise multiplication). 

\begin{figure}[h]
    \centering
    \begin{minipage}{0.48\textwidth}
        \centering
        \begin{tcolorbox}
        [colback=msblue!5!white,colframe=msblue!80!black]
        \fontsize{8pt}{8pt}\selectfont
        % \textbf{Prompt for Question Generation}\\ 
        You are a math teacher. Now, you need to help your students to learn the following math knowledge. There are some key points and example problems:

        ......

        Using these key points and example problems as a guideline, please construct a new, original math problem that requires an understanding and application of all the \textbraceleft len of selected kps\textbraceright ~knowledge points.

        ......

        Write your new math problem, using \textless Q\textgreater and \textless /Q\textgreater to indicate the question.

        \end{tcolorbox}
        \caption{Prompt for Question Generation}
        \label{fig:prompt_gen_q}
    \end{minipage}
    \hfill
    \begin{minipage}{0.5\textwidth}
        \centering
        \includegraphics[width=1.0\textwidth]{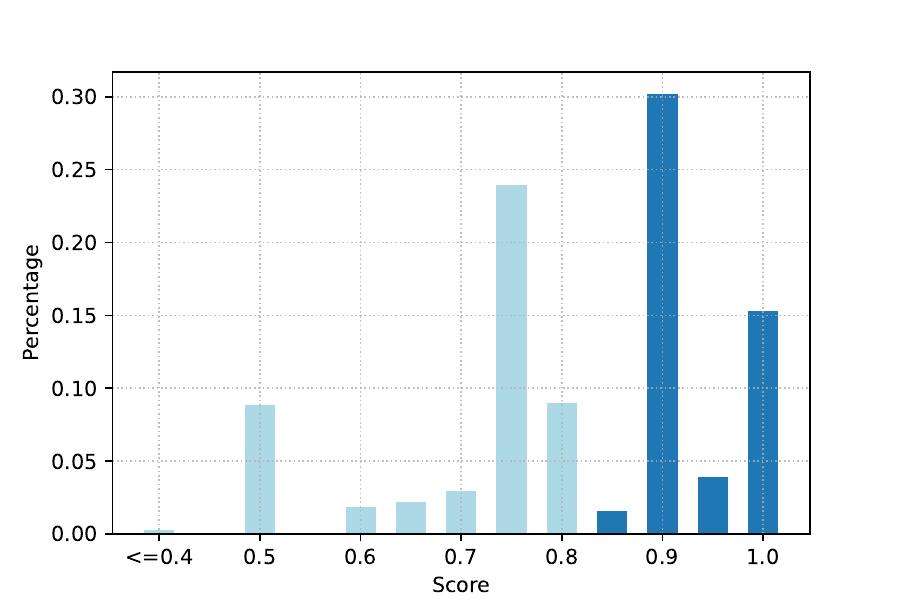}
        \caption{Score Distribution of Synthetic Questions}
        \label{fig:score_distribution}
    \end{minipage}
\end{figure}

We proceed to sample two to three topics, and for each topic, we randomly select a problem along with the associated KPs for that topic. This process yields a foundational KPs-Practice information set as the basis for our problem generation. Employing GPT-4, we use this set to generate new problems, with the prompt presented in \cref{fig:prompt_gen_q}.

Following the generation of problems, we conduct a quantitative evaluation to determine the quality of each problem by GPT-4 , prompt shown in \cref{fig:prompts}.
This assessment is based on two criteria: the presence of the provided KPs and the absence of logical or factual errors. Each problem is assigned a quality score on a continuous scale from 0 to 1. \cref{fig:score_distribution} shows the score distribution of our synthetic questions, In assembling quality-assured questions, a threshold of 0.85 is instituted to screen the newly generated problems, save about 51\% high-quality question. \cref{fig:qa_example} displays an example of a high-quality and a poor-quality problem originating from identical initial inputs.

%todo 调整caption

% \setlength{\columnsep}{0.2cm}
% \begin{figure}[ht]
% \begin{tcolorbox}[colback=msblue!5!white,colframe=msblue!80!black]
% \fontsize{8pt}{8pt}\selectfont
% % \textbf{Prompt for Question Generation}\\ 
% You are a math teacher. Now, you need to help your students to learn the following math knowledge. There are some key points and example problems:

% ......

% Using these key points and example problems as a guideline, please construct a new, original math problem that requires an understanding and application of all the \textbraceleft len of selected kps\textbraceright ~knowledge points.

% ......

% Write your new math problem, using \textless Q\textgreater and \textless /Q\textgreater to indicate the question.

% \end{tcolorbox}
% % \vspace{-1em}
% \caption{Some prompts used in this work.}
% \label{fig:prompts}
% \end{figure}

% \input{figs/new_q_example}

%\input{figs/prompt_eval_q}

\subsection{Solution Generation with Consensus Assessment}
\label{sec:answer_generation}

Prior work in the domain did not emphasize the quality control measures or relied solely on answers generated by models like GPT-4. By integrating a voting protocol, our methodology is designed to minimize the effects of noisy data and enhance the reliability of the answer-generation process. To ensure the correctness of generated answers, we employ a few-shot strategy where the reference problem is utilized as a demonstration input. To procure a diverse array of CoT rationales, we employ nucleus sampling, thereby invoking multiple prompts.
Subsequently, a voting mechanism, derived from an enhanced version of the script from \cite{gou2023tora}, is employed to aggregate the solutions. This mechanism leverages packages such as \texttt{sympy} \footnote{\url{https://www.sympy.org}} to ensure that equivalent answers, albeit in different forms (e.g., fractions and decimals), are recognized as equal.

As illustrated in \cref{fig:multi_subq}, some samples in our dataset include multiple sub-questions. We have excluded data with more than three sub-questions to maintain analytical clarity. For the multipart questions in our study, we extract the answers to sub-questions and apply a distinct voting mechanism for each. For each sub-question, we utilized GPT-4 with a temperature of 0.75 and a top-p of 0.95, resampling to obtain 10 potential responses, which then contribute to the formation of the Consensus Score Vector (CSV). Let \( x \) be a question- with \( n \) sub-questions. Then \( CSV(x) \) is defined as  
\[  
\text{CSV}(x) = [c_1, c_2, \ldots, c_n] 
\]  
where each \( c_i \) is the consensus score for the \( i \)-th sub-question and is calculated based on the voting results from the potential responses. Each \( c_i \) is in the range [0, 1].  

% \input{figs/multi_q_case}

% \subsection{Question-Answer Pair Construction}

% To augment the quantity and diversity of the data, we perform question paraphrasing on the consensus QA pairs. Hence, for each unique problem, we generate a list of paraphrased questions and a collection of answers that, while sharing the same final answer, feature distinct reasoning processes. We randomly pair these questions and answers, yielding 30 pairs for each unique problem. 

\section{Experiment}

\subsection{Training Dataset Construction}

\paragraph{KPMATH-M (252K)} This segment is based on the MATH~\citep{MATH} dataset's training set, which consists of 7,500 samples from high school math competitions, encompassing seven subjects and five difficulty levels. Utilizing the KPDDS approach on the seed problems, we generate a collection of 500K question-answer pairs. 
Considering that voting may produce multiple answers to the same question, such as in extreme cases where one question has ten answers, this type of data may not be conducive to model learning. Therefore, by rewriting each original question and its answers (not necessarily correct), we can obtain non-repetitive question-answer pairs. 
After a thorough examination of the consensus voting strategies optimization, detailed in Section \cref{sec:voted_threshold}, we refined our dataset to include the most representative 253K data points.

\paragraph{KPMATH-G (613K)} Drawing from the GSM8K~\citep{GSM8K} training set, which offers 7,473 samples of grade school math problems characterized by their 2 to 8 step solutions, we established the KPMATH-G component.
We simplified our approach due to the dataset's emphasis on basic math operations. 
Instead of generating solutions through consensus assessment, we generated three potential solutions containing mathematical expressions for each question and then meticulously verified the accuracy of each expression.
We removed any data with incorrect expressions and transformed the remaining correct solutions into detailed, expression-free explanations. This process contributed an additional 613K data points to our dataset.

\paragraph{MixMath (711K)} To ensure diversity and quality, we curated a comprehensive collection from various high-quality open-source mathematical reasoning datasets. The collection encompasses the complete datasets of MetaMath~\citep{yu2023metamath}, MMIQC~\citep{liu2024augmenting}, and Open-Platypus~\citep{platypus2023}, in addition to the training sets of GSM8K~\citep{GSM8K}, MATH~\citep{MATH}, and TAL-SCQ5K-EN~\citep{TAL-SCQ5K}, as well as the CoT subset of MathInstruct~\citep{yue2023mammoth}. As there was significant overlap among these datasets, we applied min-hash techniques to minimize redundancy. We also omitted entries with excessively long numbers or those with empty answers. This careful curation resulted in a robust dataset of 711K data points.

It is noteworthy that these procedural steps of deduplication and filtering out excessively long numbers were also applied to KPMATH-M and KPMATH-G datasets. 
Through these comprehensive measures, the final KPMATH-Plus dataset aggregates the three individual components into a substantial collection, culminating in a total of 1,576K data points that embody the richness and variety of mathematical problem-solving challenges.

\subsection{Implementation Details} In our supervised fine-tuning (SFT) experiments, we employed chat message templates to transform question-answer pairs into the format: ``User: \{question\}\textbackslash nEnclose the final answer using \textbackslash boxed\{\}.\textbackslash n\textbackslash nAssistant: \{answer\}''.
We utilized the LLaMa-Factory repository \citep{zheng2024llamafactory} to fine-tune the models for 3 epochs across all experiments.
We adopted a linear learning rate schedule with a $3\%$ warm-up ratio. The maximum learning rate is 1e-5, except for DeepSeekMath, which is 5e-5.
We trained all models with BFloat16 numerical format, \textit{DeepSpeed ZeRO Stage3} \citep{rajbhandari2021zero} and \textit{Flash-Attention 2} \citep{dao2023flashattention2}.
% The training was configured with a micro-batch size of 16 and a gradient accumulation step of 2, conducted on a single node equipped with 8$\times$ A100 GPUs.
For evaluation, we adopted the same template in SFT to prompt all questions. We employed greedy decoding with a maximum sequence length of 2,048 tokens.

\begin{table}[t]
\caption{Results on six mathematical reasoning tasks. The results of our model are bolded.
% The best results are indicated with underlined bold text, while the second-best results are bold. 
ZS: Zero-shot inference without demonstrations. Vanilla models are tested with CoT.
}
\label{tab:main}
\centering
\setlength{\tabcolsep}{4pt}
\resizebox{\linewidth}{!}{%
\begin{tabular}{lrcc|cccccc|c}
\toprule
\textbf{Model} & \textbf{Base} & \textbf{Size} & \textbf{ZS} & \textbf{GSM8k} & \textbf{MATH} & \textbf{SVAMP} & \textbf{TabMWP} & \textbf{ASDiv} & \multicolumn{1}{c|}{\textbf{MAWPS}} & \textbf{AVG} \\
% \multirow{2}{*}{\textbf{AVG}} \\
% \cmidrule{1-10}
% \multicolumn{4}{l|}{\textbf{Used as Seed Data?}} & \cmark & \cmark & \xmark & \xmark & \xmark & \xmark & \\
\midrule
\multicolumn{11}{c}{Proprietary Models} \\
\midrule
GPT-4 (0613) & - & - & \xmark &  92.0 & 42.5 & 93.1 & 67.1 & 91.3 & 97.6 & 80.6 \\
% GPT-4 (PAL) \code & - & \cmark & \xmark & 94.2 & 51.8  & 94.8 & 95.9 & 92.6 & 97.7 & 87.8 \\
% \midrule
ChatGPT & - & - & \xmark & 80.8 & 35.5  & {83.0} & {69.1} & {87.3} & {94.6} & 75.1 \\
% ChatGPT (PAL) \code & - & \cmark & \xmark & {78.6} & {38.7} & 77.8 & {79.9} & {81.0} & 89.4 & 74.2 \\
Claude-2 & - & - & \xmark & {85.2} & 32.5  & - & - & - & - & - \\
PaLM-2 & - & 540B & \xmark & 80.7 & {34.3} & - & - & - & - & - \\
 \midrule
\multicolumn{11}{c}{Open-Source Models} \\
 \midrule
Llama-2 & - & 7B & \xmark & 13.3 & 4.1  & 38.0 & 31.1 & 50.7 & 60.9 & 33.0 \\
Llama-2 SFT & - & 7B & \cmark & 41.3 & 7.2  & 31.9 & 27.8 & 47.4 & 60.0 & 35.9 \\
% Llama-2 RFT & 7B & \xmark & \cmark & 51.2 & - & - & - & - & - & -\\
Platypus-2 & Llama-2 & 7B  & \xmark & 14.4 & 5.4 & 36.7 & 26.5 & 47.9 & 58.4 & 31.6 \\
MAmmoTH & Llama-2 & 7B & \cmark & 45.9& 7.3& 48.7& 28.9& 62.3& 74.8 &44.7 \\
WizardMath & Llama-2 & 7B  & \cmark & 54.9 & 10.7 & 57.3 & 38.1 & 59.1 & 73.7 & 49.0 \\
% CodeLlama (PAL) \code & 7B & \cmark & \xmark & 34.0 & 16.6 & 59.0 & {47.3} & 61.4 & 79.6 & 49.7 \\
% Toolformer \calc & 7B & \cmark & \cmark & - & - & 29.4 & - & 40.4 & 44.0 & - \\
MetaMath & Llama-2 & 7B & \cmark & 66.6 & 20.7 & 68.8 & 43.8 & 72.5 & 86.9 & 59.9 \\
Mistral & - & 7B & \xmark & 42.9 & 12.9  & 65.1 & 55.6 & 68.4 & 86.8 & 55.3 \\
MAmmoTH & Mistral & 7B & \cmark & 52.7 & 14.5 & 54.1 & 49.1 & 64.9 & 77.5 & 52.1 \\
MMIQC & Mistral & 7B & \cmark & 74.8& 36.0& 73.1& 62.5& 81.9& 90.5 & 69.8 \\
MetaMath & Mistral & 7B & \xmark & 77.8 &29.0&78.6&64.7&81.1&93.4&70.8\\
\rowcolor{gray!20}
KPMath-Plus & Mistral & {7B}& \cmark & \textbf{82.1} & \textbf{46.8} & \textbf{76.4} & \textbf{66.4} & \textbf{86.7} & \textbf{94.2} & \textbf{75.4~\small{(+20.1)}} \\
DeepSeekMath & - & {7B}& \cmark & 63.3& 32.3& 73.2& 68.6& 82.9& 92.4 & 68.8 \\
% ds-math-instruct & ds-math & {7B}& \cmark & 82.3 & 46.7 & 83.7 & 79.1 & 90.1 & 95.7 & 79.6 \\
\rowcolor{gray!20}
KPMath-Plus & DSMath & {7B}& \cmark & \textbf{83.9} & \textbf{48.8} & \textbf{81.5} & \textbf{78.7} & \textbf{88.9} & \textbf{94.8} & \textbf{79.4~\small{(+10.6)}} \\
% ToRA & 7B & \cmark & \cmark & 68.8 & 40.1 & 68.2 & 42.4 & 73.9 & 88.8 & 63.7 \\
% ToRA-Code & 7B & \cmark & \cmark & 72.6 & 44.6 & 70.4 & 51.6 & 78.7 & 91.3 & 68.2 \\
% KPMath-Plus-Mistral-7B & 7B & \cmark & \cmark & \blue{\textbf{81.7}} & \blue{\textbf{46.1}} & \blue{\textbf{75.9}} & \blue{\textbf{67.7}} & \blue{\textbf{86.1}} & \blue{\textbf{93.8}} & \blue{\textbf{75.2~\small{(+)}}} \\
% KPMath-Plus-Mistral-7B-2 & 7B & \cmark & \cmark & \blue{\textbf{79.9}} & \blue{\textbf{46.6}} & \blue{\textbf{78.4}} & \blue{\textbf{66.6}} & \blue{\textbf{87.1}} & \blue{\textbf{93.6}} & \blue{\textbf{75.36~\small{(+)}}} \\
\midrule

Llama-2 & - & 13B & \xmark & 24.3 & 6.3 & 43.1 & 39.5 & 56.3 & 70.4 & 36.2 \\
Llama-2 SFT & - & 13B & \cmark &  51.1 & 9.2 & 46.3 & 35.8 & 58.6 & 75.0 & 42.6 \\
% Llama-2 RFT & 13B & \xmark & \cmark & 55.3 & - & - & - & - & - & -\\
Platypus-2 & Llama-2 & 13B & \xmark & 23.7 & 7.1 & 50.7 & 45.3 & 55.1 & 69.6 & 38.0 \\
MAmmoTH & Llama-2 & 13B & \cmark & 49.6& 9.9& 49.6& 40.5& 60.0& 73.4 &47.2\\
WizardMath & Llama-2 & 13B & \cmark & 63.9 & 14.0 & 64.3 & 46.7 & 65.8 & 79.7 & 51.8 \\
MetaMath & Llama-2 & 13B & \cmark &71.0& 23.2& 71.9& 52.8& 75.7& 87.0 & 63.6 \\
\rowcolor{gray!20}
KPMath-Plus & Llama-2 & 13B & \cmark & \textbf{81.6}& \textbf{41.0}& \textbf{76.7}& \textbf{63.9}& \textbf{83.2}& \textbf{92.3} & \textbf{73.1~\small{(+36.9)}} \\
% CodeLlama (PAL)  & 13B & \cmark & \xmark & 39.9 & 19.9 & 62.4 & {59.5} & 65.3 & 86.0 & 53.1 \\

% ToRA & 13B & \cmark & \cmark & {72.7} & {43.0} & {72.9} & 47.2 & {77.2} & {91.3} & {65.9} \\
% ToRA-Code & 13B & \cmark & \cmark & 
% 75.8 & 48.1 & 60.5 & 65.4 & 81.4 & 92.5 & 71.3 \\
\midrule
% Llama-1 RFT & 34B & \xmark & \cmark & 57.9 & - & - & - & - & - & -\\
% CodeLlama (PAL) \code & 34B & \cmark & \xmark & 53.3 & 23.9 & 71.0 & 63.1 & 72.4 & 91.5 & 60.7\\
Llemma & - & 34B & \xmark & 55.4& 24.4& 68.0& 57.2& 75.9& 90.5 & 61.9 \\
% MetaMath & Llemma & 34B &  \\
MMIQC & Llemma & 34B & \cmark &  79.2& {38.7}& {80.4}& 70.1& {85.0}& 94.0 &  {74.6}\\
\rowcolor{gray!20}
KPMath-Plus & Llemma & 34B & \cmark &  \textbf{82.4}& \textbf{48.6}& \textbf{81.2}& \textbf{71.9}& \textbf{87.5}& \textbf{94.5} &  \textbf{77.7~\small{(+15.8)}}\\
% ToRA-Code & 34B & \cmark & \cmark & 80.7 & 50.8 & 80.5 & 70.5 & 84.2 & 93.3 & 74.8 \\
\midrule

Llama-2 & - & 70B & \xmark & 57.8 & 14.4& 73.6 & 57.5 & 76.0 & 92.4 & 58.2 \\
Llama-2 SFT & - & 70B & \cmark & 69.3 & 14.9 & 64.0 & 53.0 & 71.3 & 84.8 & 56.6 \\
% Llama-2 RFT & 70B & \xmark & \cmark & 64.8 & - & - & - & - & - & -\\
Platypus-2 & Llama-2 & 70B & \xmark & 45.9 & 15.0 & 74.3 & 47.3 & 72.7 & 91.1 & 53.0 \\
WizardMath & Llama-2 & 70B& \cmark & {81.6} & {22.7} & {80.0} & 49.8 & {76.2} & 86.2 & {63.8} \\
MetaMath & Llama-2 & 70B & \cmark  & 82.0& 27.2& 85.8 & 63.4& 84.0& 95.4 & 73.0\\
MAmmoTH & Llama-2  & 70B & \cmark & 65.1& 14.6& 60.1& 38.2& 70.2& 80.3 & 54.8 \\
\rowcolor{gray!20}
KPMath-Plus & Llama-2 & 70B & \cmark & \textbf{87.4}& \textbf{48.6}& \textbf{81.2}& \textbf{75.1}& \textbf{89.0}& \textbf{95.4} & \textbf{79.4~\small{(+21.2)}} \\
Qwen1.5 & - & 72B & \xmark & 77.6 & 38.2& 82.5 & 52.0 & 85.1 & 95.9 & 71.9 \\
\rowcolor{gray!20}
KPMath-Plus & Qwen1.5 & 72B & \cmark &  \textbf{87.0}& \textbf{58.3}& \textbf{82.1}& \textbf{76.7}& \textbf{89.2}& \textbf{95.5} &  \textbf{81.5~\small{(+9.6)}}\\

% MMIQC & DeepSeek-LLM  & 67B & \cmark & 83.2 & 40.7 & 84.1& 74.0& 87.6& 94.9& 77.42 \\
% Llama-2 (PAL) \code & 70B & \cmark & \xmark & 55.2 & 18.3 & 74.6 & {59.5} & 71.9 & {92.8} & 60.3 \\

% ToRA & 70B & \cmark & \cmark & 84.3 & 49.7 & 82.7 & 74.0 & 86.8 & 93.8 & 76.9 \\
% \midrule
% \rowcolor{gray!20}
% KPMath-Plus & Mistral & \textbf{7B}& \cmark & \underline{\textbf{82.1}} & \underline{\textbf{46.8}} & 76.4 & \textbf{66.4} & \underline{\textbf{86.7}} & \textbf{94.2} & \underline{\textbf{75.4~\small{(+20.1)}}} \\
\bottomrule
\end{tabular}
}%
\end{table}

\subsection{Evaluation and Metrics}

We evaluate our fine-tuned models on GSM8k \citep{cobbe2021gsm8k} and MATH \citep{hendrycksmath2021}, along with 4 out-of-distribution datasets, namely SVAMP \citep{patel2021nlp_svamp}, ASDIV \citep{miao2021diverse_asdiv}, TabMWP \citep{lu2022dynamic_tabmwp}, MAWPS \citep{koncel2016mawps}.
We utilize an enhanced version of the script from \cite{gou2023tora} to extract answers, parse expressions, and compare the equivalency of the answers.
We report the zero-shot PASS@1 accuracies of predicted answers.

The Hungarian Exam was first introduced by Grok-1~\citep{xai2023}, designed to evaluate the out-of-domain capabilities of mathematical models. We follow the evaluation method proposed by \cite{finals_exam}, which divides this exam into 33 challenging problems suitable for model processing, and these answers require manual verification by humans.

\subsection{Baselines}
We present results from a range of state-of-the-art (SoTA) proprietary LLMs, including OpenAI's GPT-4~\citep{openai2023gpt4}, ChatGPT (gpt-3.5-turbo), Google's PaLM-2\citep{anil2023palm}, and Anthropic's Claude-2\citep{anthropic2023claude}. Regarding open-source models, we consider base models such as LLaMA-2\citep{Touvron2023Llama2O}, DeepSeekMath\citep{shao2024deepseekmath}, Mistral\citep{jiang2023mistral}, Llemma~\citep{azerbayev2023llemma}, and Qwen1.5\citep{bai2023qwen}. Supervised Fine-Tuning (SFT) employs CoT rationales from the original GSM8k and MATH dataset (15k samples) for fine-tuning. We also showcase the performance of advanced models using SFT or RLHF on various mathematical reasoning datasets, including MAmmoTH~\citep{yue2023mammoth}, WizardMath~\citep{luo2023wizardmath}, Platypus-2~\citep{platypus2023}, MetaMath~\citep{yu2023metamath} and MMIQC~\citep{liu2024augmenting}.

\subsection{Main Results}

Table \ref{tab:main} presents the results on six widely-used mathematical benchmarks, highlighting several key observations:
KPMath-Plus significantly enhances the performance of multiple base models, with average accuracy improvements ranging from 10.6\% to 36.9\%. The KPMath-Plus-Qwen1.5-72B model achieves zero-shot PASS@1 accuracies of 87.0\% on GSM8K and 58.3\% on MATH, and also reaches promising performance on other math reasoning datasets, outperforming competitors in the 7B to 70B range.
% Mistral-7B-KPMath-Plus demonstrates a 20.1\% performance increase over the baseline model, Mistral-7B. Among all models of the 7B parameter scale, it emerges as the leading math reasoning model, exhibiting the best performance across all datasets. Remarkably, its average performance exceeds that of all competitors within the 7B to 70B model size range, thus achieving SoTA status.

\begin{figure}[h]
    \centering
    \begin{minipage}{0.48\textwidth}
        \centering
        \includegraphics[width=1.0\textwidth]{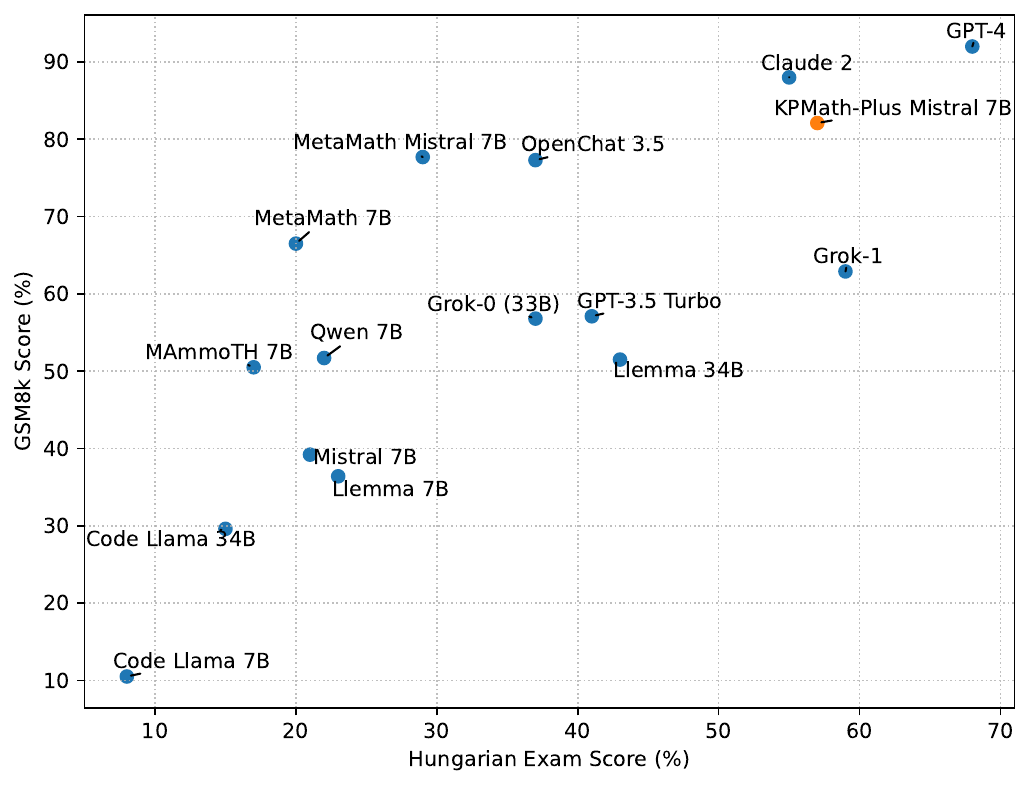}
        \caption{Hungarian Exam Score vs GSM8K Performance of Various Models.}
        \label{fig:exam_result}
    \end{minipage}
    \hfill
    \begin{minipage}{0.5\textwidth}
        \centering
        \includegraphics[width=1.0\textwidth]
        {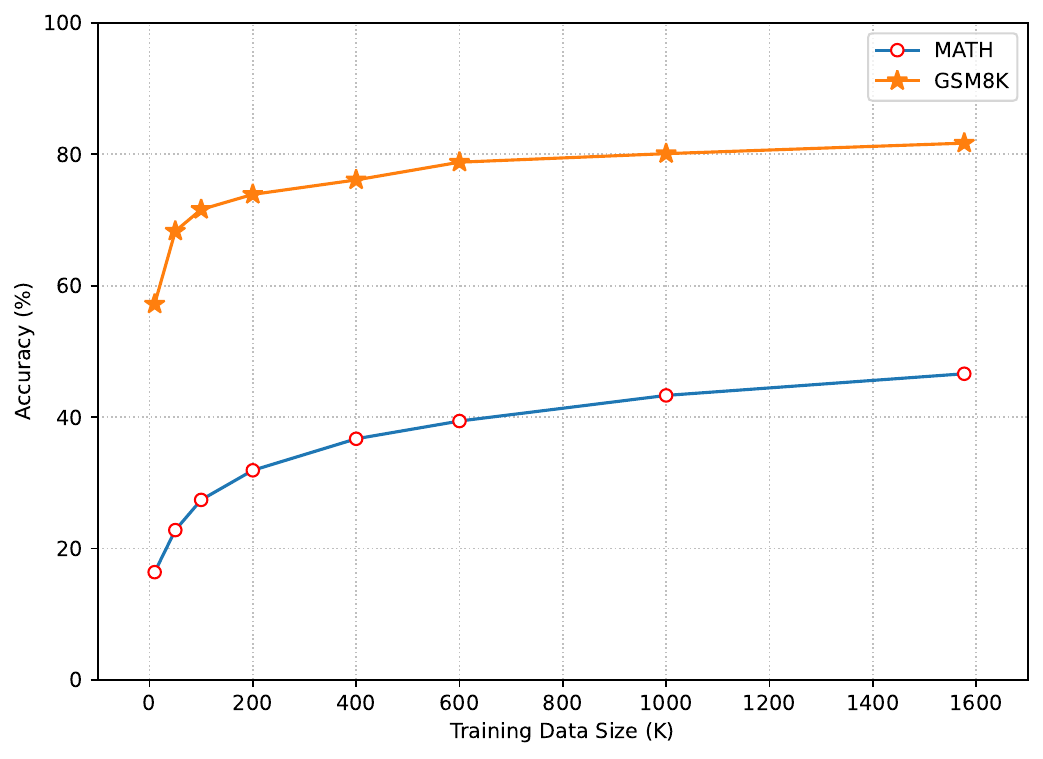}
        \caption{Performance of KPMath-Plus-Mistral-7B across various training data size.}
        \label{fig:data_size}
    \end{minipage}
\end{figure}

\cref{fig:exam_result} displays the Hungarian Exam Score versus GSM8K Performance of various models, with comparative data sourced from \cite{finals_exam}. KPMath-Plus-Mistral-7B is notably behind only to GPT-4~\citep{openai2023gpt4} and Grok-1~\citep{xai2023}. Additionally, compared to other fine-tuned models, it exhibits a well-balanced performance between the two test sets, suggesting that our model does not overfit the seed data. 

Our comprehensive analysis across multiple widely recognized math reasoning datasets confirms the superiority of KPMath-Plus in achieving the highest performance. Remarkably, KPMath-Plus maintains exceptional competitiveness even when compared to numerous 70B models, despite being based on a 7B architecture.

% \begin{figure}[h]
%     \centering
%     \begin{subfigure}{0.5\textwidth}
%         \centering
%     \includegraphics[width=1.0\textwidth]{figs/new_figs/math_gsm8k.pdf}
%         \caption{}
%         \label{fig:data_size}
%     \end{subfigure}
%     \hfill
%     \begin{subfigure}{0.48\textwidth}
%         \centering
%     \includegraphics[width=1.0\textwidth]{figs/new_figs/exam_vs_gsm8k.pdf}
%         \caption{}
%         \label{fig:ab_threshold}
%     \end{subfigure}
%     \caption{(a) Comparative analysis of fine-tuning results on Mistral-7B using various datasets, illustrating both the total number and the synthetic number of data samples. (b) Comparative analysis of the minimum acceptable score in the CSV.}
%     \label{fig:exam}
% \end{figure}

% \caption{Training Results with Different Voting Strategies (average performance on six mathematical dataset).}
% \label{fig:ab_threshold}

% {figs/new_figs/math_gsm8k.pdf}
%         \caption{Performance of KPMath-Plus-Mistral-7B across various training data size.}
%         \label{fig:data_size}

% \begin{wrapfigure}{R}{0.5\textwidth}
%     \vskip -0.05in
%     \includegraphics[width=\linewidth]{figs/new_figs/math_gsm8k.pdf}
%     \vskip -0.1in
% \caption{Performance of KPMath-Plus-Mistral-7B across various training data size..}
% \label{fig:data_size2}
% \end{wrapfigure}

\subsection{Ablation Study on Training Data Components and Size}
We conducted an ablation study with the KPMath-Plus data components on the Mistral-7B model, training over 3 epochs.
Results in \cref{tab:data_construction} indicate that integrating KPMath-G, derived from the GSM8K dataset, enhances performance on GSM8K by 5\% compared to training solely on MathMix. 
Improvements extend to SVAMP, ASDiv, and MAWPS, while a slight performance decline in MATH and TabMWP is observed, potentially due to their higher complexity.
Moreover, combining KPMath-M, based on the MATH dataset, with MixMath consistently increases scores by over 1\% across all datasets.
Merging KPMath-G and KPMath-M significantly boosts overall performance, with gains of 6.4\% on GSM8K and 3.5\% on MATH, averaging a 4.1\% improvement, illustrating the comprehensive benefits of our synthesized data within KPMath-Plus for mathematical reasoning.

We also investigated the impact of training data size on the KPMath-Plus-Mistral-7B model's performance. As demonstrated in \cref{fig:data_size}, model performance exhibits a logarithmic increase with the expansion of training data.
The model achieves impressive results with small data size and maintains a steady growth trend.
This study underlines the exceptional quality of our data and establishes a clear linkage between training data size and model performance, particularly in tackling complex tasks. In our future work, we aim to further explore larger and higher-quality datasets to continue improving model performance.

\begin{table}[h]
\caption{Performance comparison of different data components (\%).}
\label{tab:data_construction}
% \vskip 0.05in
\begin{center}
\resizebox{\columnwidth}{!}{%
% \begin{small}
\begin{tabular}{lcccccccc}
\toprule
\textbf{Data} &\textbf{GSM8K} & \textbf{MATH} & \textbf{SVAMP} & \textbf{TabMWP} & \textbf{ASDiv} & \textbf{MAWPS} & \textbf{AVG} \\
\midrule
MixMath & 75.7 & 43.3 & 73.6 & 63.1 & 82.9 & 89.1 & 71.3 \\
MixMath + KPMath-G & 80.7 & 43.0 & 76.7 & 60.0 & 85.1 & 93.9 & 73.1\\
MixMath + KPMath-M & 77.0 & 45.9 & 74.0 & 65.0 & 84.6 & 92.0 & 73.1 \\
\midrule
KPMath-Plus &\textbf{82.1}~\small{(+6.4)} &\textbf{46.8}~\small{(+3.5)}&\textbf{76.4}~\small{(+2.8)} &\textbf{66.4}~\small{(+3.3)} &\textbf{86.7}~\small{(+3.8)} &\textbf{94.2}~\small{(+5.1)} &\textbf{75.4}~\small{(+4.1)} \\

\bottomrule
\end{tabular}
% \end{small}
}
\end{center}
% \vskip -0.1in
\end{table}

\subsection{Investigation on the Consensus Voting Strategy}
\label{sec:voted_threshold}

\begin{wrapfigure}{R}{0.5\textwidth}
    \vskip -0.15in
    \includegraphics[width=\linewidth]{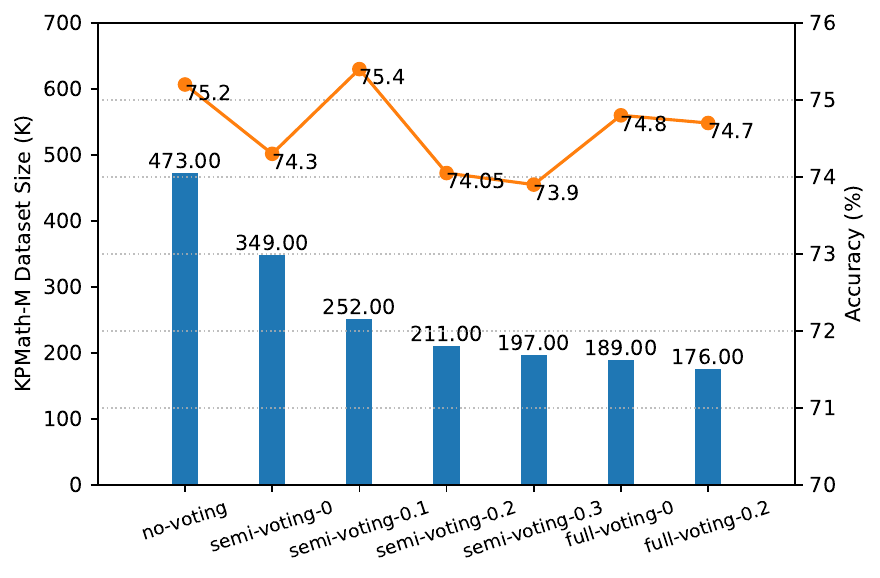}
    \vskip -0.1in
\caption{Training Results with Different Voting Strategies (average performance on six mathematical dataset).}
\label{fig:ab_threshold}
\end{wrapfigure}

We conducted a comparative analysis to identify the optimal consensus voting strategies for KPMath-M, experimenting with three distinct strategies on the Mistral-7B model. The first strategy, non-voting, involved retaining all answers, regardless of their differences.
The second strategy, semi-voting, for questions with only one sub-question, preserced only the most popular answer to ensure complete consensus in the retained response. For questions with multiple sub-questions, consensus needed to be reached on at least one of the answers.
The third strategy was full-voting, requiring consensus on every sub-question. Additionally, we conducted CSV threshold experiments on the latter two strategies.

We integrated KPMath-M with different strategies into KPMath-G and MixMath, and after fine-tuning on Mistral-7B, we obtained results as demonstrated in \cref{fig:ab_threshold}. The semi-voting with a CSV threshold of 0.1 proved to be the best setting, with the data volume reduced by 46.7\% compared to non-voting, yet without any degradation in performance. Therefore, we retained KPMath-M under this setting as our final dataset. This experiment also validated the effectiveness of our consensus voting strategy in filtering data for quality.

% \input{figs/ablation}

% As demonstrated in \cref{fig:ab_threshold}, we conducted a comparative analysis to identify the optimal minimum acceptable score within the CSV for including answers in our dataset. Our findings indicate that a $CSV_{mean}$ threshold of 0.2 is most effective in ensuring high-quality data. This threshold effectively filters out less reliable answers while retaining a robust dataset for model training.

% \subsection{Unsolvable or Wrong Questions}

% Within the KPMath dataset, specific problems are not immediately identifiable as erroneous or unsolvable but become apparent upon further examination, as exemplified in \cref{fig:wrong_q_example}. This feature is unique to our dataset relative to others concentrating exclusively on answer augmentation or question rewriting with MATH. Although these questions exhibit specific issues, the reasoning process involved remains valid. Empirical evidence demonstrates the efficacy of this subset of data's efficacy in enhancing models' mathematical capabilities, as indicated in \cref{tab:wrong_q}. Eliminating these data results in an overall performance decline of 1.1\% in the model, with a concomitant decrease observed across various mathematical subtopics.

% \input{figs/error_q_case}

% \input{tables/ab_result2}

\section{Conclusion}

In this paper, we propose a new data synthesis paradigm that is focused on the generation of large-scale, high-quality, symbolically-driven training datasets. Leveraging this paradigm, we have developed an extensive synthetic dataset tailored for mathematical reasoning. By utilizing this data set, our fine-tuned model achieved excellent performance in multiple data sets including MATH and GSM8K, and the performance exceeded all 7B to 70B competitors. Our research underscores the efficacy of integrating key points in data synthesis and applying stringent quality control protocols to both questions and answers.



% \section*{Ethics Statement}

\bibliography{ref}
\bibliographystyle{colm2024_conference}

\newpage
\appendix
\newpage
\appendix

\section{Prompts}
\label{sec:prompts}

\setlength{\columnsep}{0.2cm}
\begin{figure}[ht]
\begin{tcolorbox}[colback=msblue!5!white,colframe=msblue!80!black]
\textbf{Prompt for Knowledge Extraction}

As a mathematics education specialist, please analyze the topics and key points of the provided question and its answer. These analysis should serve as a guide for teachers to craft analogous problems and as focal learning objectives for students when approaching the problem. Be sure to avoid repetition of Key Points for clarity and conciseness. Specific requirements are as follows: 

1. Identify and categorize the main mathematical topics involved in the problem.  If knowledge from non-mathematical fields is used, it is classified into Others - xxx, such as Others - Problem Context.

2. For each topic, enumerate the essential Key Points relevant to the problem. 

------

Question: Compute $\cos 330^\circ$.
Answer: We know that $330^\circ = 360^\circ - 30^\circ$.
Since $\cos(360^\circ - \theta) = \cos \theta$ for all angles $\theta$,
we have $\cos 330^\circ = \cos 30^\circ$.
Since $\cos 30^\circ = \frac{\sqrt{3}}{2}$,
we can conclude that $\cos 330^\circ = \boxed{\frac{\sqrt{3}}{2}}$.

Analysis: \textless AN\textgreater  \textless l\textgreater Trigonometry - Cosine Function\textless /l\textgreater  \textless k\textgreater Understanding co-terminal angles in trigonometry\textless /k\textgreater  \textless k\textgreater Trigonometric identities, specifically the cosine of an angle related to a reference angle\textless /k\textgreater  \textless k\textgreater Knowledge of exact values of cosine for common angles (30°, 45°, 60°, etc.)\textless /k\textgreater  \textless k\textgreater Subtraction of angles and use of angle identities\textless /k\textgreater  \textless /AN\textgreater 

------

Question: Cara is sitting at a circular table with her five friends as shown below. How many different possible pairs of people could Cara be sitting between?

\textless asy\textgreater 
draw(circle((0,0),1));
label("$\_$",1.5dir(0));
label("$\_$",1.5dir(60));
label("Cara",1.5dir(120));
label("$\_$",1.5dir(180));
label("$\_$",1.5dir(240));
label("$\_$",1.5dir(300));
\textless \/asy\textgreater 

Answer: The number of pairs of neighbors for Cara actually has nothing to do with the shape of the table she is sitting at. That is, all that matters is that she has 5 friends and two of them will be her neighbors. There are ${5 \choose 2} = \boxed{10}$ pairs of friends that she can thus sit between.

Analysis: \textless AN\textgreater  \textless l\textgreater Combinatorics - Counting Problems\textless /l\textgreater  \textless k\textgreater Understanding of combinations and the use of the combination formula\textless /k\textgreater  \textless k\textgreater Interpreting combinatorial problems in context\textless /k\textgreater  \textless k\textgreater Application of ${n \choose k}$ to find the number of ways to choose k items from n distinct items\textless /k\textgreater  \textless /AN\textgreater  \textless AN\textgreater  \textless l\textgreater Others - Problem Context\textless /l\textgreater  \textless k\textgreater Translation of real-world scenarios into combinatorial problems\textless /k\textgreater  \textless k\textgreater Recognition that the physical arrangement (e.g., circular table) does not affect the combinatorial count\textless /k\textgreater  \textless /AN\textgreater 

------

Question: \{question of seed problem\}

Answer: \{answer of seed problem\}

Analysis:

% \Sepline
% \textbf{Prompt for Question Rephrasing}\ 
% Below is an original mathematics problem with its corresponding solution. Your task is to creatively reframe this problem into 10 different scenarios. The scenarios should incorporate the same numerical values but be set in a wide array of environments, involving various characters or objects. Ensure that each scenario stands out by using a variety of sentence structures, settings, and applications, ranging from the everyday to the fantastical. The rephrased problems should each be unique in their presentation while still clearly corresponding to the original problem's solution.

% Original Problem: \textless Q\textgreater ...... \textless /Q\textgreater

% Original Solution: \textless A\textgreater ...... \textless /A\textgreater

% Now, create the 10 rephrased questions based on this information. Using \textless Q\textgreater and \textless /Q\textgreater to indicate the question.
% Question1:

\end{tcolorbox}
% \vspace{-1em}
\caption{Prompts for knowledge extraction.}
\label{fig:prompt_knowledge}
\end{figure}

\setlength{\columnsep}{0.2cm}
\begin{figure}[ht]
\begin{tcolorbox}[colback=msblue!5!white,colframe=msblue!80!black]
% \textbf{Prompt for KPs Generation:}\\ 
% As a mathematics education specialist, please analyze the topics and key points of the provided question and its answer. These analysis should serve as a guide for teachers to craft analogous problems and as focal learning objectives for students when approaching the problem. Be sure to avoid repetition of Key Points for clarity and conciseness. Specific requirements are as follows: 

% 1. Identify and categorize the main mathematical topics involved in the problem.  If knowledge from non-mathematical fields is used, it is classified into Others - xxx, such as Others - Problem Context.

% 2. For each topic, enumerate the essential Key Points relevant to the problem. 

% \Sepline
\textbf{Prompt for Question Generation}\\ 
You are a math teacher. Now, you need to help your students to learn the following math key points. There are some key points and example problems:

......

Using these key points and example problems as a guideline, please construct a new, original math problem that requires an understanding and application of all the \textbraceleft len of selected kps\textbraceright ~key points.

......

Write your new math problem, using \textless Q\textgreater and \textless /Q\textgreater to indicate the question.

\Sepline
\textbf{Prompt for Question Evaluation}\\ 
Given these math key points and example problems:

......

I have formulated a new math problem as follows:

......

Could you please evaluate whether the new math problem incorporates all the provided key points? 
And please determine whether there are factual or logical errors in the problem.
Provide a score as a floating-point number between 0 and 1, where 1 means all key points are fully integrated into the problem and there are no factual or logical errors, and 0 indicates none are integrated or there are many factual or logical errors.
The closer the score is to 1, the more comprehensive the problem is in terms of covering the given concepts and theorems, and the fewer factual or logical errors there are.
Evaluation Score:
\Sepline
\textbf{Prompts for Question Answering}\\ 
\vspace{-1em}
\begin{itemize}[leftmargin=0.5cm, rightmargin=0.2cm]
    \setlength{\itemsep}{-0.2em}
    \item You have extensive experience with math competitions and you need to write a detailed tutorial for beginners.
    \item You are an experienced math teacher and you need to help beginners learn these math problems.
    \item There are some challenging math problems, take deep breath and solve them.
    \item Look at the following math problems and help me to solve them.
    \item You are a ai assistant know a lot of math problems. Tell me how to solve the following math problems.
\end{itemize}

\Sepline
\textbf{Prompt for Question Rephrasing}\\ 
Below is an original mathematics problem with its corresponding solution. Your task is to creatively reframe this problem into 10 different scenarios. The scenarios should incorporate the same numerical values but be set in a wide array of environments, involving various characters or objects. Ensure that each scenario stands out by using a variety of sentence structures, settings, and applications, ranging from the everyday to the fantastical. The rephrased problems should each be unique in their presentation while still clearly corresponding to the original problem's solution.

Original Problem: \textless Q\textgreater ...... \textless /Q\textgreater

Original Solution: \textless A\textgreater ...... \textless /A\textgreater

Now, create the 10 rephrased questions based on this information. Using \textless Q\textgreater and \textless /Q\textgreater to indicate the question.
Question1:

\end{tcolorbox}
% \vspace{-1em}
\caption{Some prompts used in this work.}
\label{fig:prompts}
\end{figure}

\clearpage
\section{Mathematic Topic Details}

\label{sec:appd-topics}

\begin{table*}[h]
\caption{Topics in MATH.}
\label{tab:topics_math}
\vskip 0.15in
\begin{center}
\begin{small}
\resizebox{1.0\textwidth}{!}{
% Type scores: {'Algebra': 56.9, 'Counting & Probability': 33.8, 'Geometry': 30.3, 'Intermediate Algebra': 19.5, 'Number Theory': 31.1, 'Prealgebra': 58.9, 'Precalculus': 23.4}
\begin{tabular}{rlrlrl}
\toprule
1 & Calculus - Functions and their Properties & 2 & Calculus - Optimization & 3 & Calculus - Limits \\
4 & Algebra - Solving Equations & 5 & Algebra - Polynomials & 6 & Algebra - Inequalities \\
7 & Algebra - Functions & 8 & Algebra - Simplifying Expressions & 9 & Algebra - Linear Equations \\
10 & Algebra - Quadratic Equations & 11 & Algebra - Square Roots & 12 & Algebra - Radicals \\
13 & Algebra - Sequences and Series & 14 & Algebra - Linear Functions & 15 & Algebra - Complex Numbers \\
16 & Algebra - Function Operations & 17 & Algebra - Exponents & 18 & Algebra - Rational Functions \\
19 & Algebra - Function Transformations & 20 & Algebra - Proportions & 21 & Algebra - Proportional Relationships \\
22 & Algebra - Logarithms & 23 & Algebra - Substitution & 24 & Algebra - Exponential Growth \\
25 & Algebra - Summation & 26 & Algebra - Absolute Value & 27 & Algebra - Variables and Expressions \\
28 & Algebra - Ratios and Proportions & 29 & Algebra - Geometric Series & 30 & Algebra - Interval Notation \\
31 & Algebra - Polynomial Expansion & 32 & Algebra - Real Numbers & 33 & Others - Problem Context \\
34 & Others - Graph Interpretation & 35 & Others - Problem Solving & 36 & Arithmetic - Order of Operations \\
37 & Arithmetic - Time Calculations & 38 & Arithmetic - Division & 39 & Arithmetic - Basic Operations \\
40 & Arithmetic - Fractions & 41 & Arithmetic - Multiplication & 42 & Arithmetic - Percentages \\
43 & Arithmetic - Addition & 44 & Arithmetic - Averages & 45 & Arithmetic - Rate Problems \\
46 & Arithmetic - Unit Conversion & 47 & Arithmetic - Rounding Numbers & 48 & Number Theory - Fractions and Decimals \\
49 & Number Theory - Integer Properties & 50 & Number Theory - Powers and Roots & 51 & Number Theory - Floor Function \\
52 & Number Theory - Floor and Ceiling Functions & 53 & Number Theory - Perfect Squares & 54 & Number Theory - Divisibility \\
55 & Number Theory - Factors and Multiples & 56 & Number Theory - Prime Numbers & 57 & Number Theory - Multiples \\
58 & Number Theory - Odd and Even Numbers & 59 & Number Theory - Digit Sums & 60 & Number Theory - Modulo Arithmetic \\
61 & Number Theory - Properties of Integers & 62 & Number Theory - Units Digit & 63 & Number Theory - Greatest Common Divisor (GCD) \\
64 & Number Theory - Perfect Squares and Cubes & 65 & Number Theory - Counting Digits & 66 & Number Theory - Modular Arithmetic \\
67 & Number Theory - Division and Remainders & 68 & Number Theory - Powers and Exponents & 69 & Geometry - Circles \\
70 & Geometry - Coordinate Geometry & 71 & Geometry - Distance Formula & 72 & Geometry - Polygons \\
73 & Geometry - Midpoint Formula & 74 & Geometry - Reflections & 75 & Geometry - Area Calculation \\
76 & Geometry - Lines and Angles & 77 & Geometry - Perimeter & 78 & Geometry - Parabolas \\
79 & Geometry - Area of a Circle & 80 & Geometry - Rectangles & 81 & Geometry - Triangles \\
82 & Geometry - Transformations & 83 & Geometry - Squares & 84 & Geometry - 3D Shapes \\
85 & Geometry - Angles & 86 & Geometry - Volume of Solids & 87 & Geometry - Pyramids \\
88 & Geometry - Similar Triangles & 89 & Geometry - Cones & 90 & Geometry - Parallelograms \\
91 & Geometry - Conic Sections & 92 & Geometry - Ellipse & 93 & Geometry - Coordinate Systems \\
94 & Geometry - Planes in Three Dimensions & 95 & Financial Mathematics - Compound Interest & 96 & Sequences and Series - Infinite Series \\
97 & Complex Numbers - Absolute Value & 98 & Combinatorics - Counting Problems & 99 & Combinatorics - Factorials \\
100 & Combinatorics - Binomial Coefficients & 101 & Combinatorics - Pascal's Triangle & 102 & Measurement - Unit Conversion \\
103 & Statistics - Mean & 104 & Statistics - Mean and Median & 105 & Probability - Basic Concepts \\
106 & Probability - Expected Value & 107 & Probability - Geometric Probability & 108 & Data Interpretation - Bar Graphs \\
109 & Trigonometry - Tangent Function & 110 & Trigonometry - Sine and Cosine Functions & 111 & Trigonometry - Polar Coordinates \\
112 & Set Theory - Overlapping Sets & 113 & Number Systems - Base Conversion & 114 & Number Systems - Binary Numbers \\
115 & Linear Algebra - Matrices & 116 & Linear Algebra - Vectors & 117 & Linear Algebra - Determinants \\
118 & Linear Algebra - Vectors and Parametric Equations & 119 & Vector Algebra - Dot Product \\

\bottomrule
\end{tabular}}
\end{small}
\end{center}
\vskip -0.1in
\end{table*}

\begin{table*}[h]
\caption{Topics in GSM.}
\label{tab:topics_gsm}
\vskip 0.15in
\begin{center}
\begin{small}
\resizebox{1.0\textwidth}{!}{
% Type scores: {'Algebra': 56.9, 'Counting & Probability': 33.8, 'Geometry': 30.3, 'Intermediate Algebra': 19.5, 'Number Theory': 31.1, 'Prealgebra': 58.9, 'Precalculus': 23.4}
\begin{tabular}{rlrlrl}
\toprule
1 & Arithmetic - Basic Operations & 2 & Others - Problem Context & 3 & Arithmetic - Multiplication and Division \\
4 & Arithmetic - Multiplication and Addition & 5 & Algebra - Word Problems & 6 & Word Problems - Problem Solving \\
7 & Arithmetic - Division & 8 & Arithmetic - Addition and Subtraction & 9 & Arithmetic - Percentages \\
10 & Arithmetic - Fractions & 11 & Proportional Reasoning & 12 & Arithmetic - Time Calculations \\
13 & Arithmetic - Averages & 14 & Geometry - Volume Calculation & 15 & Measurement - Length \\
16 & Arithmetic - Sequences and Series & 17 & Geometry - Rectangles & 18 & Arithmetic - Comparison \\
19 & Set Theory - Overlapping Sets & 20 & Arithmetic - Rate Problems & 21 & Arithmetic - Unit Conversion \\
22 & Arithmetic - Money & 23 & Geometry - Triangles & 24 & Problem Solving - Multi-step Problems \\
25 & Probability & 26 & Geometry - Area Calculation & 27 & Arithmetic - Age Problems \\
28 & Geometry - Perimeter & 29 & Measurement - Volume & 30 & Word Problems - Distance Problems \\
31 & Measurement - Weight & 32 & Arithmetic - Subtraction and Multiplication & 33 & Algebra - Exponential Growth \\
34 &Fractions - Addition and Subtraction \\
\bottomrule
\end{tabular}}
\end{small}
\end{center}
\vskip -0.1in
\end{table*}

\begin{figure}[h]
\centering
\includegraphics[width=0.98\textwidth]{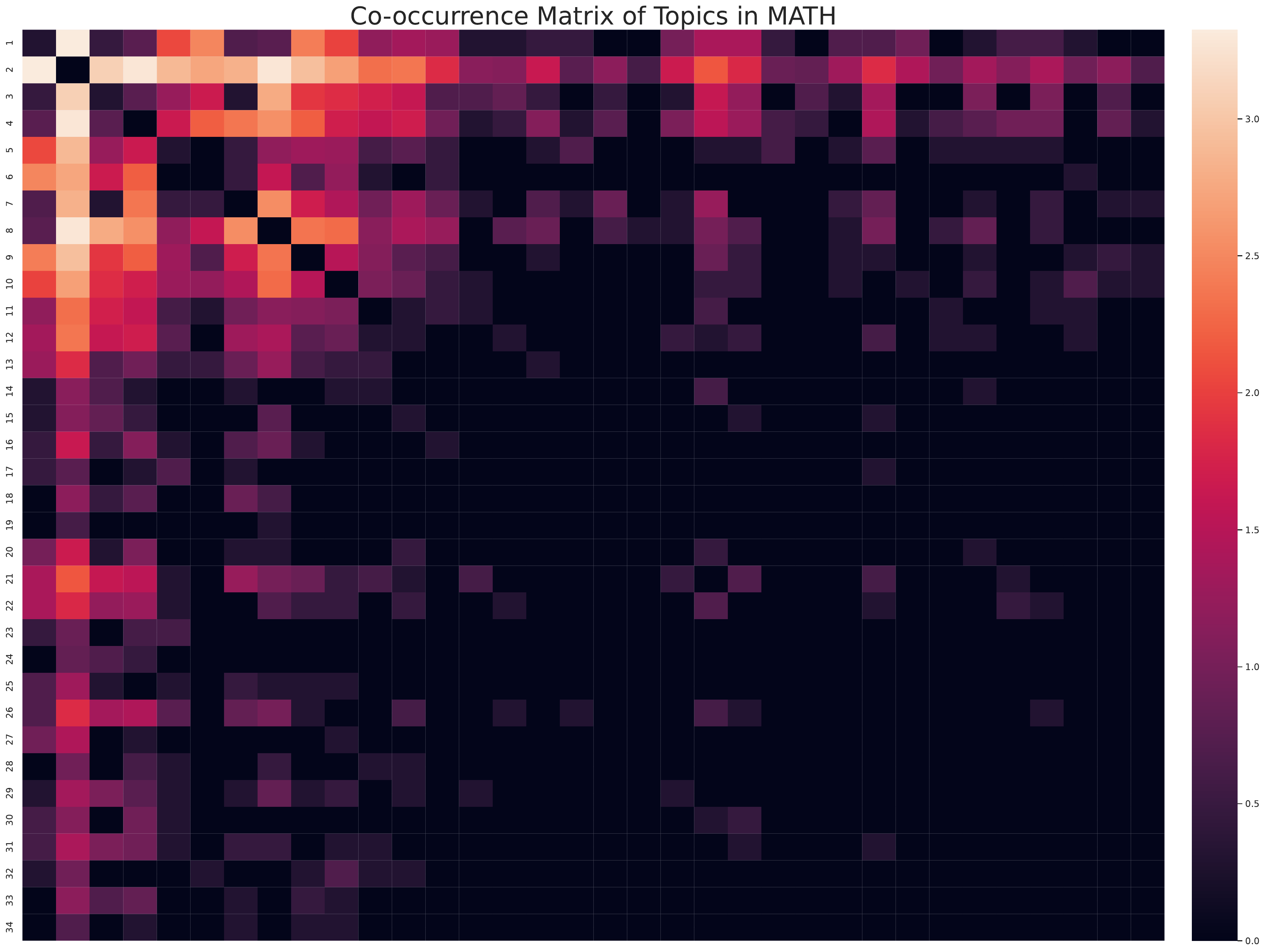}
\caption{Visualized heat map of co-occurrence probability matrix.}
\label{fig:co-occurrence_matrix_gsm}
\end{figure}

\begin{figure}[h]
\centering
\includegraphics[width=0.98\textwidth]{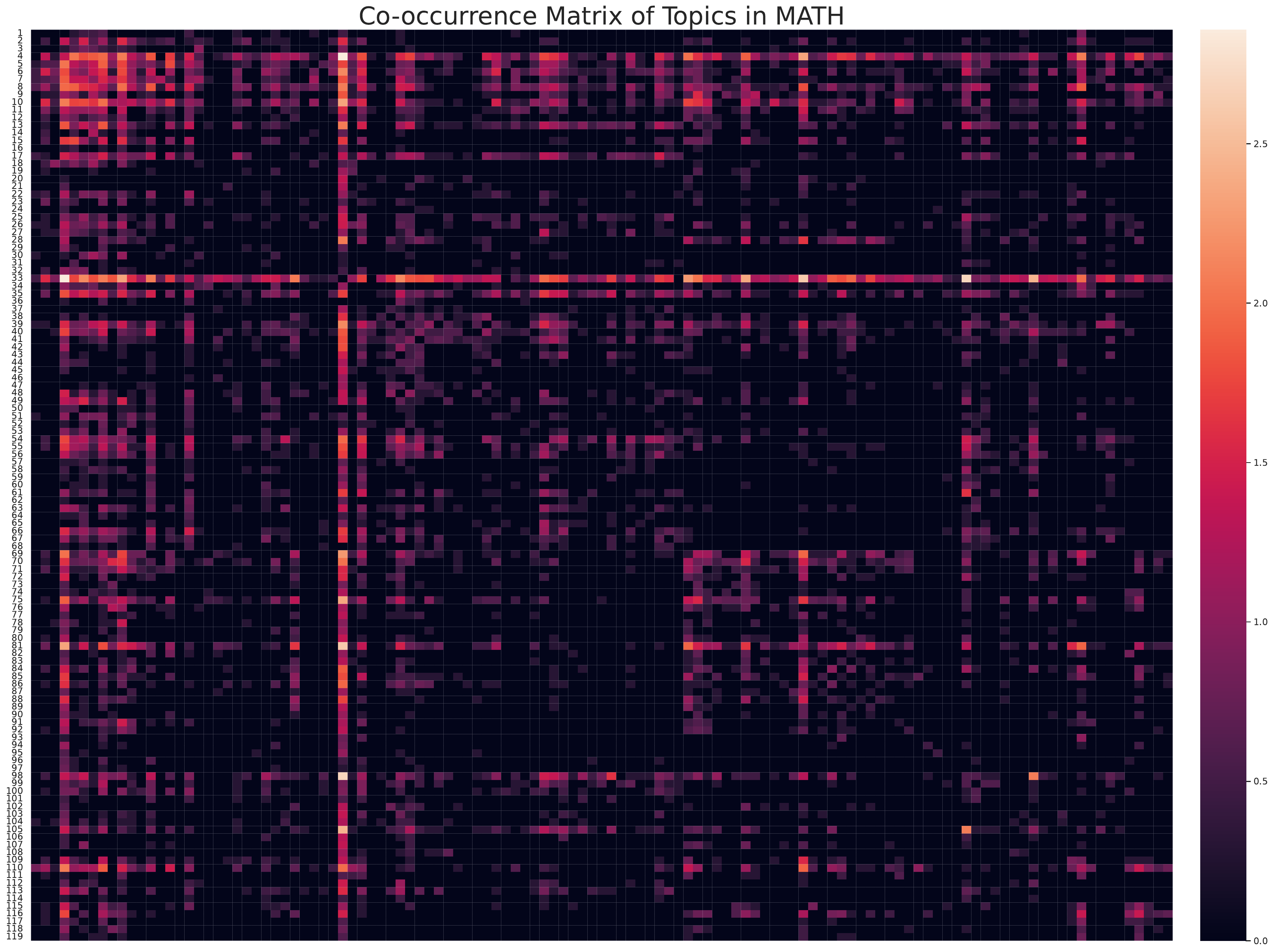}
\caption{Visualized heat map of co-occurrence probability matrix.}
\label{fig:co-occurrence_matrix_math}
\end{figure}

\clearpage
\section{QA Examples}

\begin{figure}[h]

\begin{minipage}{1\textwidth}
\begin{tcolorbox}[colback=msyellow!5!white,colframe=msyellow!80!black, fontupper=\small]
\textbf{Synthetic Question Bad Case ~\xmark}

You are a treasure hunter and have found a chest full of gold coins in an ancient ruin. The chest contains less than 1000 gold coins. You decided to distribute the coins equally among your 13 fellow treasure hunters. However, after dividing your gold coins into 13 equal piles, you realize that if you give away all your gold coins, 3 people will receive an extra gold coin. You also found an ancient riddle in the chest stating that the sum of the digits of the number of gold coins is a base 13 number that leaves the same remainder upon division by 10 as the number itself. What is the largest number of gold coins you could have to cause this to happen?
\end{tcolorbox}

\begin{tcolorbox}[colback=msyellow!5!white,colframe=msyellow!80!black, fontupper=\small]
\textbf{Synthetic Question Good Case ~\cmark}

You have a number of books that you were planning to distribute equally among your 15 classmates. However, after dividing your books into 15 equal piles, you realize that if you give away all your books, 3 classmates will receive an extra book. You have less than 130 books. Meanwhile, a friend of yours asked you the base 8 sum of the digits of the number of books you have. What is this base 8 sum?

% \Sepline
% \textbf{Voted Solution for the Good Case}: \\
% You have a number of books that you were planning to distribute equally among your 15 classmates. However, after dividing your books into 15 equal piles, you realize that if you give away all your books, 3 classmates will receive an extra book. You have less than 130 books. Meanwhile, a friend of yours asked you the base 8 sum of the digits of the number of books you have. What is this base 8 sum?", "solution": "If the three classmates received one fewer book, then the number of books you would have would be a multiple of 15. However, there are three extra books there, so the number of books you have can be written in the form $15k+3$. We have that $15k+3 < 130$, so $k < \frac{127}{15}$. Since $k$ is the number of books each person is receiving, $k$ must be an integer, so we have that $k = 8$. Therefore, the largest number of books you could have is $15(8) + 3 = 123$.

% But this number has to be expressed in base 8. The number $123$ in base 10 is equal to $173$ in base 8. Thus, the base 8 sum of the digits of the number of books you have is $1+7+3=\boxed{11}$. The answer is: 11.

\end{tcolorbox}
\end{minipage}
% \centering
% \begin{tikzpicture}
% \draw [line width=1mm, -latex] (0,0) -- (0,-0.5);
% \end{tikzpicture}

% \begin{minipage}{1\textwidth}
% \begin{tcolorbox}[colback=msyellow!5!white,colframe=msyellow!80!black, fontupper=\small]
% \textbf{Voted Solution for the Good Case}: \\
% If the three classmates received one fewer book, then the number of books you would have would be a multiple of 15. However, there are three extra books there, so the number of books you have can be written in the form $15k+3$. We have that $15k+3 < 130$, so $k < \frac{127}{15}$. Since $k$ is the number of books each person is receiving, $k$ must be an integer, so we have that $k = 8$. Therefore, the largest number of books you could have is $15(8) + 3 = 123$.

% But this number has to be expressed in base 8. The number $123$ in base 10 is equal to $173$ in base 8. Thus, the base 8 sum of the digits of the number of books you have is $1+7+3=\boxed{11}$. The answer is: 11.

% \end{tcolorbox}
% \end{minipage}

% \vspace{-1em}
\caption{Examples of high-quality and poor-quality problems generated from the same initial inputs.}
\label{fig:qa_example}
\end{figure}

\begin{figure}[h]
\begin{minipage}{1\textwidth}
\begin{tcolorbox}[colback=msyellow!5!white,colframe=msyellow!80!black, fontupper=\small]
\textbf{Question}:
Tommy ran four blocks south and then 9 blocks east for his morning run. If each block is one-fourth of a mile, how many miles did he run? After his run, he relaxed by watching the sunrise. He noticed that the sun rose at 6:00 AM and by 6:40 AM, it had risen 30 degrees in the sky. If the sun continues to rise at this rate, what will be the angle of the sun above the horizon at 7:20 AM?
\Sepline
\textbf{Solution}: Tommy ran $4+9=13$ blocks. This is $13\left(\frac{1}{4}\right)=\boxed{3.25}$ miles. 

For the second part of the question, we need to determine the rate at which the sun is rising. We know that in 40 minutes (from 6:00 AM to 6:40 AM), the sun rose 30 degrees. So, the rate of rise is $\frac{30\text{ degrees}}{40\text{ minutes}} = \frac{3}{4}$ degrees per minute. From 6:40 AM to 7:20 AM is another 40 minutes. So in this time, the sun will rise an additional $40\text{ minutes} \times \frac{3}{4}\text{ degrees/minute} = 30$ degrees.
Therefore, at 7:20 AM, the sun will be $30\text{ degrees (from 6:00 AM to 6:40 AM)} + 30\text{ degrees (from 6:40 AM to 7:20 AM)} = \boxed{60}$ degrees above the horizon. The answer is: 3.25; 60
\end{tcolorbox}
\end{minipage}
\caption{A complex problem comprising multiple sub-questions in KPMath.}
\label{fig:multi_subq}
\end{figure}

\clearpage
\section{Data Contamination Test}
\label{sec:contamination}

To mitigate the risk of data contamination in our evaluation benchmark, we adhere to the methodology presented by \cite{azerbayev2023llemma} for cross-examining $n$-gram overlaps between our synthetic dataset and the test sets of Math and GSM8K. A hit is recorded if any $n$-gram in the test sequence appears within any training set. Given the shorter length of questions, we implement a 20-gram check for questions and a 30-gram check for solutions. For GSM8K, our analysis identifies no hits. For Math, our analysis identifies 102 hits for KPMath questions and 108 hits for KPMath solutions, fewer than the 181 and 144 hits found in the MATH training set's problems and solutions, respectively. Notably, KPMath accounts for 9 unique problem hits, and 16 solution hits absent in the MATH train set, with details provided in \cref{sec:contamination}. Moreover, we conducted a manual review of all hits. We determined that they were instances of repeated problem contexts or intermediate reasoning steps rather than exact duplicates of questions or solutions. This examination indicates a very low risk of data contamination for KPMath.

\begin{lstlisting}[caption={Caces of Question 20-gram Hits}, label={listing:contamination_q}]
Question of Math Test 257:
A square is drawn such that one of its sides coincides with the line $y = 7$, and so that the endpoints of this side lie on the parabola $y = 2x^2 + 8x + 4$. What is the area of the square?
Overlap: 
A square is drawn such that one of its sides coincides with the line $y = 7$, and so that the endpoints of this side lie on the parabola $y =
>>>>>>>>

Question of Math Test 2314:
Let $x,$ $y,$ and $z$ be nonnegative real numbers such that $x + y + z = 3.$ Find the maximum value of \[(xy + z)(xz + y).\]
Overlap: 
$x,$ $y,$ and $z$ be nonnegative real numbers such that $x + y + z = 3.$ Find the maximum value of
>>>>>>>>

Question of Math Test 2253:
For some real numbers $a$ and $b$, the equation \[ 8x^3 + 4ax^2 + 2bx + a = 0 \]has three distinct positive roots. If the sum of the base-2 logarithms of the roots is 5, what is the value of $a$?
Overlap: 
three distinct positive roots. If the sum of the base-2 logarithms of the roots is 5, what is the value of
>>>>>>>>

Question of Math Test 4974:
Find the matrix $\mathbf{M}$ that swaps the rows of a matrix. In other words, \[\mathbf{M} \begin{pmatrix} a & b \\ c & d \end{pmatrix} = \begin{pmatrix} c & d \\ a & b \end{pmatrix}.\]If no such matrix $\mathbf{M}$ exists, then enter the zero matrix.
Overlap: 
\begin{pmatrix} a & b \\ c & d \end{pmatrix} = \begin{pmatrix} c & d \\ a & b \end{pmatrix}.\]If no such matrix
>>>>>>>>

Question of Math Test 2416:
Let $a,$ $b,$ $c,$ and $d$ be positive real numbers such that $a + b + c + d = 10.$ Find the maximum value of $ab^2 c^3 d^4.$
Overlap: 
Let $a,$ $b,$ $c,$ and $d$ be positive real numbers such that $a + b + c + d =
>>>>>>>>

Question of Math Test 1042:
Suppose the function $f(x)$ is defined explicitly by the table $$\begin{array}{c || c | c | c | c | c} x & 0 & 1 & 2 & 3 & 4 \\ \hline f(x) & 0 & 0 & 1 & 3 & 6 \end{array}$$ This function is defined only for the values of $x$ listed in the table. Suppose $g(x)$ is defined as $f(x)-x$ for all numbers $x$ in the domain of $f.$ How many distinct numbers are in the range of $g(x)?$
Overlap: 
$$\begin{array}{c || c | c | c | c | c} x & 0 & 1 & 2 & 3 & 4 \\ \hline
>>>>>>>>

Question of Math Test 1972:
In the diagram, the two triangles shown have parallel bases. What is the ratio of the area of the smaller triangle to the area of the larger triangle? Express your answer as a common fraction. [asy] path p = (0,0)--dir(-60)--dir(-120)--(0,0); draw(p); draw(scale(3)*p); label("4 cm",dir(-60)--dir(-120),S); label("10 cm",3*dir(-60)--3dir(-120),S); [/asy]
Overlap: 
What is the ratio of the area of the smaller triangle to the area of the larger triangle? Express your answer as a common fraction. [asy]
>>>>>>>>

Question of Math Test 1917:
Points $A(0,0), B(9,6)$ and $C(6,12)$ are vertices of triangle $ABC$. Point $D$ is on segment $AB$ such that $2(AD) = DB$, point $E$ is on segment $BC$ such that $2(BE) = EC$ and point $F$ is on segment $CA$ such that $2(CF) = FA$. What is the ratio of the area of triangle $DEF$ to the area of triangle $ABC$? Express your answer as a common fraction.
Overlap: 
= FA$. What is the ratio of the area of triangle $DEF$ to the area of triangle $ABC$? Express your answer as a common fraction.
>>>>>>>>

Question of Math Test 3166:
The greatest common divisor of two integers is $(x+3)$ and their least common multiple is $x(x+3)$, where $x$ is a positive integer. If one of the integers is 40, what is the smallest possible value of the other one?
Overlap: 
integers is $(x+3)$ and their least common multiple is $x(x+3)$, where $x$ is a positive integer. If one of the integers is 40, what is the smallest possible value of
>>>>>>>>
\end{lstlisting}
\bigskip
\begin{lstlisting}[caption={Caces of Solution 30-gram Hits}, label={listing:contamination_s}]
Solution of Math Test 3232:
Let us look at the powers of $5$: \begin{align*} 5^1 &\equiv 5 \pmod{7} \\ 5^2 &\equiv 4 \pmod{7} \\ 5^3 &\equiv 6 \pmod{7} \\ 5^4 &\equiv 2 \pmod{7} \\ 5^5 &\equiv 3 \pmod{7} \\ 5^6 &\equiv 1 \pmod{7}. \end{align*} Since $5^6 \equiv 1 \pmod{7},$ we see that $5^{30} \equiv (5^6)^5 \equiv 1 \pmod{7},$ hence our desired remainder is $\boxed{1}.$
Overlap: 
\begin{align*} 5^1 &\equiv 5 \pmod{7} \\ 5^2 &\equiv 4 \pmod{7} \\ 5^3 &\equiv 6 \pmod{7} \\ 5^4 &\equiv 2 \pmod{7} \\ 5^5 &\equiv 3 \pmod{7} \\ 5^6 &\equiv 1 \pmod{7}.
>>>>>>>>

Solution of Math Test 1445:
We can just treat the knot as another bead. There are $5!$ ways to place the beads and the knot on the bracelet, but we must divide by 5 for rotational symmetry (5 rotations for each arrangement), and by 2 for reflectional symmetry (we can flip the bracelet to get the same arrangement). The answer is $\dfrac{5!}{5 \times 2} = \boxed{12}$.
Overlap: 
on the bracelet, but we must divide by 5 for rotational symmetry (5 rotations for each arrangement), and by 2 for reflectional symmetry (we can flip the bracelet to get the same arrangement). The answer is $\dfrac{5!}{5 \times 2} =
>>>>>>>>

Solution of Math Test 3528:
The primes between 1 and 100 are 2, 3, 5, 7, 11, 13, 17, 19, 23, 29, 31, 37, 41, 43, 47, 53, 59, 61, 67, 71, 73, 79, 83, 89, and 97. We compute their residues modulo 16: 2, 3, 5, 7, 11, 13, 1, 3, 7, 13, 15, 5, 9, 11, 15, 5, 11, 13, 3, 7, 9, 15, 3, 9, 1. We multiply all of these numbers modulo 16, taking advantage of the fact that $3\cdot 5 \equiv -1 (\text{mod }16)$, $7\cdot9\equiv -1 (\text{mod }16)$, $11\cdot 13\equiv -1 (\text{mod }16)$, and $15\equiv -1(\text{mod }16)$. We find that our answer is $\boxed{6}$.
Overlap: 
between 1 and 100 are 2, 3, 5, 7, 11, 13, 17, 19, 23, 29, 31, 37, 41, 43, 47, 53, 59, 61, 67, 71, 73, 79, 83, 89, and 97.
>>>>>>>>

Solution of Math Test 458:
The sum of the first $n$ odd integers is $1 + 3 + \dots + (2n - 1)$. The sum of an arithmetic series is equal to the average of the first and last term, multiplied by the number of terms, so this sum is $[1 + (2n - 1)]/2 \cdot n = n^2$. Then the sum of the odd integers between 0 and 100 is $50^2$, and the sum of the odd integers between 0 and 200 is $100^2$. Therefore, the ratio of the sum of the odd integers between 0 and 100 to the sum of the odd integers between 100 and 200 is $\frac{50^2}{100^2-50^2}=\frac{1}{4-1}=\boxed{\frac{1}{3}}$.
Overlap: 
+ \dots + (2n - 1)$. The sum of an arithmetic series is equal to the average of the first and last term, multiplied by the number of terms, so
>>>>>>>>

Solution of Math Test 3562:
Let's find the cycle of units digits of $7^n$, starting with $n=1$ : $7, 9, 3, 1, 7, 9, 3, 1,\ldots$ . The cycle of units digits of $7^{n}$ is 4 digits long: 7, 9, 3, 1. Thus, to find the units digit of $7^n$ for any positive $n$, we must find the remainder, $R$, when $n$ is divided by 4 ($R=1$ corresponds to the units digit 7, $R=2$ corresponds to the units digit 9, etc.) Since $53\div4=13R1$, the units digit of $7^{53}$ is $\boxed{7}$.
Overlap: 
$7, 9, 3, 1, 7, 9, 3, 1,\ldots$ . The cycle of units digits of $7^{n}$ is 4 digits long: 7, 9, 3, 1. Thus, to find the units digit of $7^n$ for any positive $n$, we must find the remainder, $R$, when $n$ is divided by 4 ($R=1$ corresponds to the units digit 7, $R=2$ corresponds to the units digit 9, etc.) Since
>>>>>>>>

Solution of Math Test 2349:
Note that \[ab + bc + cd \le ab + bc + cd + da = (a + c)(b + d).\]By AM-GM, \[(a + c)(b + d) \le \left( \frac{(a + c) + (b + d)}{2} \right)^2 = \frac{1}{4}.\]Equality occurs when $a = 0,$ $b = \frac{1}{2},$ $c = \frac{1}{2},$ and $d = 0,$ so the maximum value of $ab + bc + cd$ is $\boxed{\frac{1}{4}}.$
Overlap: 
that \[ab + bc + cd \le ab + bc + cd + da = (a + c)(b + d).\]By AM-GM, \[(a + c)(b + d) \le \left( \frac{(a + c) + (b + d)}{2} \right)^2 =
>>>>>>>>

Solution of Math Test 333:
Let $d$ be the common difference. Then the last term is $7 + (15-1)d = 7+14d$. The sum of an arithmetic series is equal to the average of the first and last term, multiplied by the number of terms, so the sum of the series is \[\frac{7 + (7 + 14d)}{2} \cdot 15 = 15(7d + 7) = 105d + 105.\]We are told that this sum equals $-210$, so we have $105+105d = -210$, from which we find $d=\boxed{-3}$. Note: $\boxed{3}$ is also accepted as an answer.
Overlap: 
The sum of an arithmetic series is equal to the average of the first and last term, multiplied by the number of terms, so the sum of the series is
>>>>>>>>

Solution of Math Test 3695:
The measure of each interior angle in a regular $n$-gon is $180(n-2)/n$ degrees. Therefore, the measure of angle $\angle BAD$ is $180(6-2)/6=120$ degrees and the measure of angle $CAD$ is 108 degrees. Their difference, $\angle BAC$, measures $120-108=\boxed{12\text{ degrees}}$.
Overlap: 
The measure of each interior angle in a regular $n$-gon is $180(n-2)/n$ degrees. Therefore, the measure of angle $\angle BAD$ is $180(6-2)/6=120$ degrees and the measure of angle $CAD$ is
>>>>>>>>

Solution of Math Test 79:
The radius of the circular path of the horse closer to the center is $\frac{1}{4}$ of the radius of the path of the horse farther from the center. Since circumference is directly proportional to radius, the length of shorter path is $\frac{1}{4}$ of the length of the longer path. Therefore, 4 times as many revolutions must be made to go the same distance, which is $27\times4=\boxed{108}$ revolutions.
Overlap: 
radius of the circular path of the horse closer to the center is $\frac{1}{4}$ of the radius of the path of the horse farther from the center. Since circumference is directly proportional to radius, the length of
>>>>>>>>

Solution of Math Test 593:
Moving terms to the LHS, we have $x^2-6x+y^2+8y=24$. Completing the square on the quadratic in $x$, we add $(6/2)^2=9$ to both sides. Completing the square on the quadratic in $y$, we add $(8/2)^2=16$ to both sides. We are left with the equation $x^2-6x+9+y^2+8y+16=49 \Rightarrow (x-3)^2+(y+4)^2=49$. Thus, our circle has center $(3,-4)$. The distance between this center and the point $(-3,-12)$ is $\sqrt{(-3-3)^2+(-12-(-4))^2}=\boxed{10}$.
Overlap: 
Completing the square on the quadratic in $x$, we add $(6/2)^2=9$ to both sides. Completing the square on the quadratic in $y$, we add $(8/2)^2=16$ to both sides. We are left with the equation
>>>>>>>>

Solution of Math Test 4816:
By the product-to-sum identities, we have that $2\cos a \sin b = \sin (a+b) - \sin (a-b)$. Therefore, this reduces to a telescoping series: \begin{align*} \sum_{k=1}^{n} 2\cos(k^2a)\sin(ka) &= \sum_{k=1}^{n} [\sin(k(k+1)a) - \sin((k-1)ka)]\\ &= -\sin(0) + \sin(2a)- \sin(2a) + \sin(6a) - \cdots - \sin((n-1)na) + \sin(n(n+1)a)\\ &= -\sin(0) + \sin(n(n+1)a)\\ &= \sin(n(n+1)a). \end{align*}Thus, we need $\sin \left(\frac{n(n+1)\pi}{2008}\right)$ to be an integer; this integer can be only $\{-1,0,1\}$, which occurs when $2 \cdot \frac{n(n+1)}{2008}$ is an integer. Thus $1004 = 2^2 \cdot 251$ divides $n(n+1)$. Since 251 is prime, 251 must divide $n$ or $n + 1.$ The smallest such $n$ is 250, but 1004 does not divide $250 \cdot 251.$ The next smallest such $n$ is 251, and 1004 divides $251 \cdot 252.$ Therefore, the smallest such integer $n$ is $\boxed{251}.$
Overlap: 
we have that $2\cos a \sin b = \sin (a+b) - \sin (a-b)$. Therefore, this reduces to a telescoping series: \begin{align*} \sum_{k=1}^{n} 2\cos(k^2a)\sin(ka) &= \sum_{k=1}^{n} [\sin(k(k+1)a) - \sin((k-1)ka)]\\ &= -\sin(0) + \sin(2a)- \sin(2a) + \sin(6a) - \cdots - \sin((n-1)na) + \sin(n(n+1)a)\\ &= -\sin(0) +
>>>>>>>>

Solution of Math Test 2959:
Let $x = \frac{a}{b},$ $y = \frac{b}{c},$ and $z = \frac{c}{a}.$ Then $x + y + z = 7$ and $\frac{1}{x} + \frac{1}{y} + \frac{1}{z} = 9.$ Also, \[xyz = \frac{a}{b} \cdot \frac{b}{c} \cdot \frac{c}{a} = 1,\]so $xy + xz + yz = 9.$ We want to compute $x^3 + y^3 + z^3.$ Recall the factorization \[x^3 + y^3 + z^3 - 3xyz = (x + y + z)(x^2 + y^2 + z^2 - xy - xz - yz).\]Squaring the equation $x + y + z = 7,$ we get \[x^2 + y^2 + z^2 + 2(xy + xz + yz) = 49.\]Then \[x^2 + y^2 + z^2 - xy - xz - yz = 49 - 3(xy + xz + yz) = 49 - 3 \cdot 9 = 22.\]Hence, \[x^3 + y^3 + z^3 = 7 \cdot 22 + 3 = \boxed{157}.\]
Overlap: 
the factorization \[x^3 + y^3 + z^3 - 3xyz = (x + y + z)(x^2 + y^2 + z^2 - xy - xz - yz).\]Squaring the equation $x + y + z =
>>>>>>>>

Solution of Math Test 1237:
We consider the subset $\{ 2, 3, 5, 7, 11 \}$ which consists only of the prime integers in the original set. Any subset consisting entirely of prime numbers must be a subset of this particular subset. And, there are $2^5 - 1 = \boxed{31}$ non-empty subsets of this 5-element set, which we can easily see by making the choice of including or not including each element.
Overlap: 
numbers must be a subset of this particular subset. And, there are $2^5 - 1 = \boxed{31}$ non-empty subsets of this 5-element set, which we can easily see by making the choice of including or not including each element.
>>>>>>>>

Solution of Math Test 503:
The function is defined when the value inside the square root is positive, i.e. we must have $x^2-5x+6>0$. Factoring, we get $(x-3)(x-2)>0$. So either both factors in the left hand side are negative or they are both positive. They are both negative when $x<2$. They are both positive when $x>3$. So the domain of $f(x)$ is $x<2 \text{ or } x>3$, or $x \in \boxed{(-\infty, 2) \cup (3, \infty)}$ in interval notation.
Overlap: 
So either both factors in the left hand side are negative or they are both positive. They are both negative when $x<2$. They are both positive when $x>3$. So the domain of
>>>>>>>>

Solution of Math Test 4954:
Let $\mathbf{v} = \begin{pmatrix} 3 \\ 6 \\ 15 \end{pmatrix}$ and $\mathbf{w} = \begin{pmatrix} 2 \\ 1 \\ -2 \end{pmatrix}.$ [asy] import three; size(180); currentprojection = perspective(6,3,2); triple I = (1,0,0), J = (0,1,0), K = (0,0,1), O = (0,0,0); triple V = (3,2,2), W = (4,1,3), P = dot(V,W)/abs(W)^2*W, R = 2*P - V; draw(V--R,dashed); draw(0.85*P--(0.85*P + 0.15*(V - P))--(P + 0.15*(V - P))); draw(O--V,Arrow3(6)); draw(P--W,Arrow3(6)); draw(O--P,Arrow3(6)); draw(O--R,Arrow3(6)); draw(O--3*I, Arrow3(6)); draw(O--3*J, Arrow3(6)); draw(O--3*K, Arrow3(6)); label("$x$", 3.2*I); label("$y$", 3.2*J); label("$z$", 3.2*K); label("$\mathbf{v}$", V, NE); label("$\mathbf{w}$", W, N); label("$\mathbf{p}$", P, SW); label("$\mathbf{r}$", R, SW); [/asy] Let $\mathbf{p}$ be the projection of $\mathbf{v}$ onto $\mathbf{w},$ so \[\mathbf{p} = \frac{\mathbf{v} \cdot \mathbf{w}}{\mathbf{w} \cdot \mathbf{w}} \mathbf{w} = \frac{\begin{pmatrix} 3 \\ 6 \\ 15 \end{pmatrix} \cdot \begin{pmatrix} 2 \\ 1 \\ -2 \end{pmatrix}}{\begin{pmatrix} 2 \\ 1 \\ -2 \end{pmatrix} \cdot \begin{pmatrix} 2 \\ 1 \\ -2 \end{pmatrix}} \begin{pmatrix} 2 \\ 1 \\ -2 \end{pmatrix} = \frac{-18}{9} \begin{pmatrix} 2 \\ 1 \\ -2 \end{pmatrix} = \begin{pmatrix} -4 \\ -2 \\ 4 \end{pmatrix}.\]Let $\mathbf{r}$ be the reflection of $\mathbf{v}$ across line $L.$ Then $\mathbf{p}$ is the midpoint of $\mathbf{v}$ and $\mathbf{r},$ so \[\mathbf{p} = \frac{\mathbf{v} + \mathbf{r}}{2}.\]Then \[\mathbf{r} = 2 \mathbf{p} - \mathbf{v} = 2 \begin{pmatrix} -4 \\ -2 \\ 4 \end{pmatrix} - \begin{pmatrix} 3 \\ 6 \\ 15 \end{pmatrix} = \begin{pmatrix} -11 \\ -10 \\ -7 \end{pmatrix}.\]Hence, the resulting point is $\boxed{(-11,-10,-7)}.$
Overlap: 
\end{pmatrix} \cdot \begin{pmatrix} 2 \\ 1 \\ -2 \end{pmatrix}}{\begin{pmatrix} 2 \\ 1 \\ -2 \end{pmatrix} \cdot \begin{pmatrix} 2 \\ 1 \\ -2 \end{pmatrix}} \begin{pmatrix} 2 \\ 1 \\ -2 \end{pmatrix} =
>>>>>>>>

Solution of Math Test 701:
We see that $x^2 + 4x + 4 = (x + 2)^2$. If $x$ must be positive, we can see that this expression can take on the value of any perfect square that is greater than or equal to $(1+2)^2=9$. The possible values between 10 and 50 are thus 16, 25, 36, and 49, achieved when $x=2,3,4,5$ respectively. So, there are $\boxed{4}$ positive integers $x$ for which $x^2+4x+4$ is between 10 and 50.
Overlap: 
see that $x^2 + 4x + 4 = (x + 2)^2$. If $x$ must be positive, we can see that this expression can take on the value of any perfect square that is greater than or equal to
>>>>>>>>
\end{lstlisting}

\end{document}